\documentclass[10pt,twocolumn,letterpaper]{article}
\usepackage[pagenumbers]{wacv}
\usepackage{graphicx}
\usepackage{amsmath}
\usepackage{amssymb}
\usepackage{booktabs}
\usepackage[pagebackref,breaklinks,colorlinks]{hyperref}
\usepackage{textcomp}
\usepackage{stfloats}
\usepackage{hyperref} 
\usepackage{url}
\usepackage{verbatim}
\usepackage{cite}
\usepackage[capitalize]{cleveref}
\crefname{section}{Sec.}{Secs.}
\Crefname{section}{Section}{Sections}
\Crefname{table}{Table}{Tables}
\crefname{table}{Tab.}{Tabs.}
\usepackage{booktabs}
\usepackage{multirow}
\usepackage[normalem]{ulem}
\usepackage{float}
\usepackage{wrapfig}
\usepackage{amsmath}
\usepackage{amssymb}
\usepackage{bbding}
\usepackage{caption}
\usepackage{algorithm}
\usepackage{algpseudocode}
\usepackage{cleveref}
\usepackage{xcolor}

\begin{document}
\title{Enhancing Zero-Shot Facial Expression Recognition by LLM Knowledge Transfer}

\author{Zengqun Zhao, Yu Cao, Shaogang Gong, Ioannis Patras\\
Queen Mary University of London\\
{\tt\small \{zengqun.zhao, yu.cao, s.gong, i.patras\}@qmul.ac.uk}
}

\maketitle

\begin{abstract}
Current facial expression recognition (FER) models are often designed in a supervised learning manner and thus are constrained by the lack of large-scale facial expression images with high-quality annotations. Consequently, these models often fail to generalize well, performing poorly on unseen images in inference. Vision-language-based zero-shot models demonstrate a promising potential for addressing such challenges. However, these models lack task-specific knowledge and therefore are not optimized for the nuances of recognizing facial expressions. To bridge this gap, this work proposes a novel method, Exp-CLIP, to enhance zero-shot FER by transferring the task knowledge from large language models (LLMs). Specifically, based on the pre-trained vision-language encoders, we incorporate a projection head designed to map the initial joint vision-language space into a space that captures representations of facial actions. To train this projection head for subsequent zero-shot predictions, we propose to align the projected visual representations with task-specific semantic meanings derived from the LLM encoder, and the text instruction-based strategy is employed to customize the LLM knowledge. Given unlabelled facial data and efficient training of the projection head, Exp-CLIP achieves superior zero-shot results to the CLIP models and several other large vision-language models (LVLMs) on seven in-the-wild FER datasets. \textit{The code is available at \url{https://github.com/zengqunzhao/Exp-CLIP}.}
\end{abstract}

\vspace{-1em}
\section{Introduction}
Facial expressions are crucial for human communication, allowing individuals to convey a wide range of information non-verbally. In the field of Human Computer Interaction, Facial Expression Recognition (FER) often aims at the analysis of the underlying emotions or sentiments, and has various applications such as assisting people who are blind or have low vision  \cite{joseph2021facial}, monitoring patients for better mental health assessment \cite{bishay2019schinet} and improving user experience when interacting with applications and devices \cite{maat2007gaze}. FER can be categorized into two types based on input distinctions: Static FER (SFER) and Dynamic FER (DFER). SFER involves analyzing individual images or snapshots of faces while DFER focuses on analyzing facial expressions across a sequence of images or frames.

\begin{figure}[!t]
	\centering
	\includegraphics[scale=0.29]{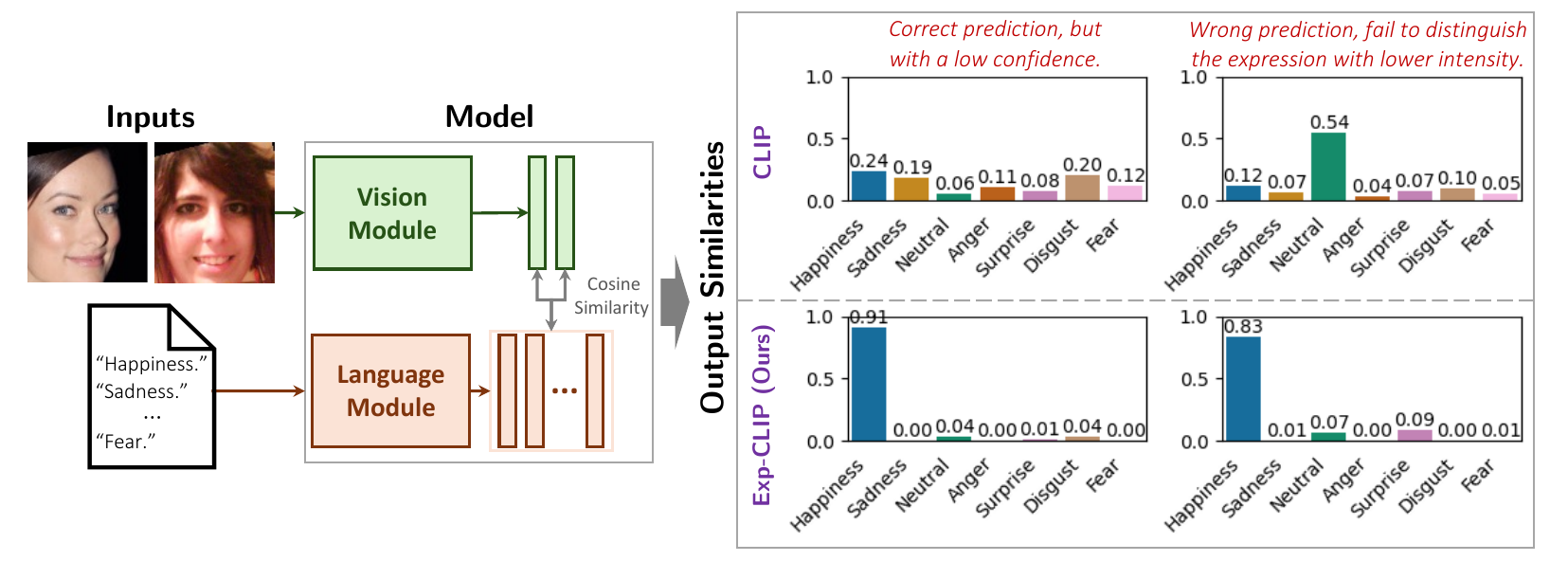}
        \vspace{-0.5em}
	\caption{CLIP model learned more general feature representations, lacking task-specific knowledge and therefore are not optimized for the nuances of recognizing facial expressions.}
	\label{fig_1}
\vspace{-2em}
\end{figure}

In the past decade, research in both SFER and DFER has shifted from controlled laboratory environments to more challenging in-the-wild conditions \cite{li2020deep}. Given the availability of large-scale facial expression datasets captured in real-world scenarios \cite{li2018reliable,m2017affectnet,jiang2020dfew,jiang2020dfew} and advancements in foundational models \cite{he2016deep,dosovitskiy2020image}, current FER models have attained high accuracy \cite{zhang2022learn,zhang2023weakly,li2023intensity,wang2023rethinking}. However, these deep FER models were mainly trained in a supervised manner, relying heavily on large-scale training data with categorical annotations. The reliance on such labelled data presents significant challenges. Firstly, the process of acquiring and annotating large-scale training data is both labour-intensive and costly. Secondly, supervised FER models tend to struggle with generalization across different domains \cite{chen2021cross}. This means that while these models may perform well within the domain they were trained on, their performance often degrades when applied to new, unseen domains. Additionally, FER annotations frequently encounter challenges related to noise and inconsistent annotations caused by subjectivity, hindering the training process \cite{wang2020suppressing,wu2023net}. Given these challenges, exploring alternative approaches becomes crucial. One promising direction is zero-shot learning, which uses semantic information to bridge the gap between seen and unseen classes. This approach is capable of generalizing to unseen expressions and removes the need for extensive annotation of training data \cite{pourpanah2022review}.

\begin{figure}[!t]
	\centering
	\includegraphics[scale=0.59]{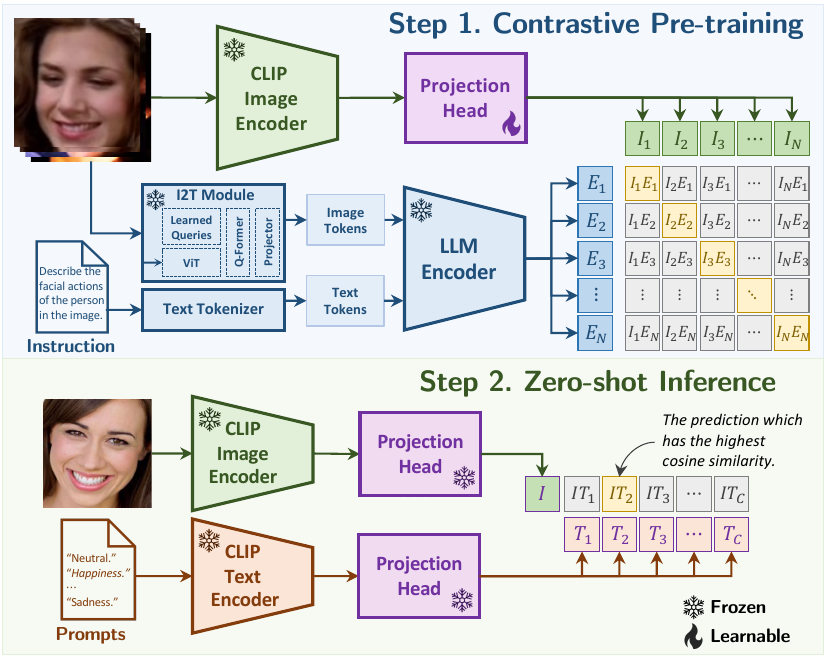}
        \vspace{-2em}
	\caption{The proposed framework in this paper introduces contrastive pre-training and zero-shot FER. In the testing phase, a learned projection head is employed to enhance image-text representations for facial expressions. This projection head is learned in an unsupervised manner, leveraging the knowledge from LLMs. The I2T module consists of a ViT, a Q-Former \cite{li2023blip}, and a projector, which is adopted to map the images into LLM tokens.}
	\label{fig_2}
\vspace{-2em}
\end{figure}

Recently, large vision-language pre-training models have shown potential for achieving promising zero-shot performance \cite{radford2021learning,zheng2022general,wu2023gpt4vis,li2023blip}. Those models are often designed using one of the following two strategies: (a) aligning large connections of images and raw texts using two separate encoders \cite{radford2021learning,zheng2022general,singh2022flava} or (b) incorporating visual modality to Large Language Models (LLMs) to create Large Vision-Language Models (LVLMs) \cite{li2023blip,li2024llava,alayrac2022flamingo}. Under the previous training paradigm, the CLIP model \cite{radford2021learning} was designed to learn from vast datasets containing millions of image-text pairs. This extensive training enables the model to achieve remarkable zero-shot capabilities, effectively categorizing images into various classes without specific training on those categories. Building on the foundations of the CLIP model, the FaRL model \cite{zheng2022general} extends this approach by focusing specifically on face-image-text pairs. This adaptation allows FaRL to develop a universal facial representation effective across a broad spectrum of face analysis tasks. Compared to previous models, FaRL demonstrates enhanced performance in accuracy and generalization. EmotionCLIP \cite{zhang2023learning} is proposed to extract visual emotion representations from verbal and nonverbal communication. However, the original CLIP model learned more general feature representations may not be optimized for the nuances of recognizing facial expressions (see Fig. \ref{fig_1}). Training texts in FaRL and EmotionCLIP are either general appearance descriptions or conversations, lacking detailed facial action of expression descriptions. Fine-tuning the CLIP model with facial expression visual data and their corresponding text descriptions, as discussed in \cite{foteinopoulou2023emoclip}, appears to be a more direct approach. However, this idea relies on facial images with detailed facial action descriptions and has the same manual labour costs as the supervised approaches. Whilst LVLM-based approaches appear to promise automation, prompting such models for zero-shot tasks demands substantial engineering efforts \cite{gu2023systematic}. Additionally, these models suffer problems caused by hallucinations of LLM decoders\cite{rawte2023survey}. It remains a significant challenge to achieve reliable zero-shot FER.

In this work, we present a novel approach to enhancing zero-shot FER model learning by leveraging LLMs knowledge transfer. Specifically, to improve the zero-shot capability of large pre-trained vision-language models for a given task, we propose to conduct projection on latent visual and textual representations to emphasize crucial task-specific elements while disregarding irrelevant ones. To this end, a shared projection head on top of the aligned image-encoder and text-encoder is introduced, serving the purpose of projecting the initial joint vision-language space into a task-specific joint space. To learn this projection head subject to optimising the downstream zero-shot FER prediction task, we consider aligning the projected visual embeddings of the image-encoder with task-specific semantic embeddings of the LLM encoder based on a contrastive training framework. By doing so, a text instruction-based strategy is utilized to customize the LLM knowledge. This alignment is beneficial to the image encoder by absorbing task-specific knowledge from LLMs. The framework of the proposed method is shown in Fig. \ref{fig_2}. Our method involves contrastive pre-training and zero-shot inference in two parts: First, contrastive pre-training is introduced for training the projection head in an unsupervised manner, and second, the learned projection head is applied to both image and text encoder in inference for the zero-shot FER task. Our contributions are:
\vspace{-1.5em}
\begin{itemize}
\itemsep -0.05cm
\item We propose to align a general vision-language feature space into a task-aware LLM feature space by learning a lightweight projection head so to improve zero-shot FER model learning based on pre-trained vision-language models. The learnable projection head is optimized through contrastive learning on unlabeled facial data in an unsupervised fashion.  
\item We utilize a text instruction-based strategy to customize the LLMs knowledge for a downstream task and elicit the task-specific semantic features of LLM. 
\item The proposed method outperforms the various LVLMs on both image-based and video-based FER datasets. 
\end{itemize}

\section{Related Work}
\subsection{Facial Expression Recognition}
\textbf{Traditional FER Models:} The traditional FER model relies on a feature extractor and a classifier. In the past decade, research on static FER has primarily focused on addressing real-world challenges such as occlusion \cite{wang2020region,zhao2021learning}, pose variation \cite{wang2019identity}, domain shift \cite{chen2021cross}, and label noise \cite{wang2020suppressing,wu2023net,zhao2021robust}. In the latest work, Wu \etal \cite{wu2023net} introduce a landmark-aware net to address the ambiguity of expressions by leveraging facial landmarks to mitigate the impact of label noise. Lee \etal \cite{lee2023latent} proposed LatentOFER to improve FER accuracy under face occlusions. In their method, the model can detect occlusions, restore occluded parts of the face as if they were unoccluded, and recognize them. To address the imbalanced FER problem, Zhang \etal \cite{zhang2024leave} propose to leverage re-balanced attention maps to regularize the model, enabling it to extract transformation invariant information about the minor classes from all training samples. In contrast, research on Dynamic FER primarily aims to develop models capable of generating robust spatial-temporal representations of facial expressions. The introduction of the Transformer to DFER, as seen in Former-DFER \cite{zhao2021former}, marked a significant advancement, demonstrating superior performance compared to methods based on 3DCNNs or CNN-RNNs. Subsequent studies have further enhanced performance by addressing issues such as noisy frames \cite{li2022nr} and the intensity of varied frames \cite{li2023intensity}. Moreover, Wang \etal \cite{wang2023rethinking} introduced M3DFEL to tackle the challenge of imbalances between short- and long-term temporal relationships in DFER. The concept of soft labelling \cite{kawamura2024midas} has also been applied to reduce data ambiguity.

\textbf{Vision-language-based FER Models:} In contrast to traditional FER models, vision-language-based models do not rely on classifiers. Instead, they derive final classification results by comparing the similarity between visual features and diverse text features, selecting the highest similarity as the ultimate prediction. Based on the CLIP, Li \etal \cite{li2023cliper} introduced CLIPER, which incorporates multiple expression text descriptors to learn fine-grained expression representations, achieving state-of-the-art performance on various FER benchmarks. Zhao \etal \cite{zhao2023dferclip} proposed DFER-CLIP, designed for dynamic FER with temporal learning, introducing textual descriptions of facial behaviour for contrastive learning. Zhang \etal \cite{zhang2023weakly} presented the CLEF model, which conducts text-driven contrastive learning for facial behaviour understanding, utilizing activity descriptions and coarse-grained information from certain datasets. However, the reliance on extensive labelled facial data in these models presents a challenge in data acquisition. The recent work by Foteinopoulou \etal \cite{foteinopoulou2023emoclip} introduces a novel vision-language model utilizing sample-level text descriptions as natural language supervision for facial expression video to enhance zero-shot FER classification. Nevertheless, this approach faces the difficulty of labelling detailed facial action descriptions, requiring expertise in perceiving nuanced facial movements. In contrast, our method utilizes facial images without annotations in an unsupervised fashion, presenting a more labour-efficient approach.

\subsection{Large Vision-Language Models}
The current landscape of LVLMs can be broadly categorized into two main types. The first type involves aligning large collections of images and raw texts by utilizing two distinct encoders \cite{radford2021learning,xu2021videoclip,ni2022expanding,zheng2022general,singh2022flava,zhang2023learning}. The second type integrates the visual modality into LLMs to form comprehensive LVLMs \cite{alayrac2022flamingo,li2023blip,li2024llava,xenos2024vllms}. 
In the context of the first type, the pioneer work CLIP \cite{radford2021learning} maps data of different modalities, text and images, into a shared embedding space. This shared space exhibits the capability to generalize to various image classification tasks with zero- and few-shot learning. Subsequently, VideoCLIP \cite{xu2021videoclip} and XCLIP \cite{ni2022expanding} were introduced for zero-shot video-text understanding. FaRL \cite{zheng2022general} focuses on learning universal facial representations from face-image-text pairs, while EmotionCLIP \cite{zhang2023learning} is designed to extract visual emotion representations from both verbal and nonverbal communication. The recent FLAVA \cite{singh2022flava} learned a foundational language and vision representation that enables unimodal vision and language understanding as well as multimodal reasoning, all within a single pre-trained model. Regarding the LLMs-based models, Flamingo \cite{alayrac2022flamingo} bridges the powerful pre-trained vision-only and language-only models, and BLIP2 \cite{li2023blip} further improves the capacity of the LVLMs by bridging the modality gap with a lightweight Querying Transformer and adopting more powerful LLM. The recent model LLaVa \cite{li2024llava} was trained by fine-tuning LLMs on GPT-generated multimodal instruction-following data, demonstrating impressive multimodel chat abilities. Nevertheless, the zero-shot FER performance of these models is challenged by substantial engineering efforts on prompting \cite{gu2023systematic} and the hallucination issue inherited from LLMs decoders\cite{rawte2023survey}.

\section{Methods}
The core idea of our method is to map a general visual representation into a task-aware representation beneficial to facial expression recognition. A projection-based strategy is proposed to fulfil this purpose, in which a projection head is learned through leveraging the task-specific knowledge from the LLMs. We will first present an overview of our method in \cref{sec:overview} and then illustrate the zero-shot FER task in \cref{sec:zeroshotfer}, finally, we will show how to train the learnable projection head for zero-shot learning in \cref{sec:knowledgetrans}.   

\subsection{Overview}
\label{sec:overview}
Fig. \ref{fig_1} shows an overview of our approach. As we can see, there are two phases involved in our method: contrastive pre-training and zero-shot inference. For zero-shot inference, the network tries to recognize the unseen facial images or videos into one of the categorical emotions. Different from the supervised approaches, the network is not trained with any labelled facial images or videos. In contrast to the plain vision-language pre-training model, we applied a shared projection head on top of both the image encoder and text encoder. The projection head is learned to project the general joint vision and language space to a task-aware joint vision and language space, boosting the zero-shot ability in the specific task. Regarding the contrastive pre-training stage, the unlabelled facial images are used as input for both the CLIP image encoder and the I2T module. The I2T module contains a ViT, a Q-Former, and a projector from BLIP2 \cite{li2023blip}, which is adapted to project the image representations into language tokens, and these tokens will feed into the LLM encoder along with the tokenized instructions. During the training, the outputs of the projection head and LLM encoder will be aligned by the contrastive loss, in which only the projection head will be optimised. The pseudo-code of the proposed method is shown in Algorithm \ref{a1}. 

\begin{algorithm}[!t]
\footnotesize
\caption{Pseudo Code of the Proposed Method}
\label{a1}
\begin{algorithmic}
\State \textbf{Require:} Pre-training Dataset $X$, Instruction $c_s$, Prompt List $C_p$, CLIP Image Encoder $E_I(\cdot)$, CLIP Text Encoder $E_T(\cdot)$, I2T Module $Q(\cdot)$, LLM-Encoder (T5) $G(\cdot)$, Text Tokenizer $T(\cdot)$
\State \textbf{Notation}: Sample $x \sim X$, Prompt $c_p^i \sim C_p$, Latent code $\textbf{\textit{e}}_I \sim E_I(x)$, $\textbf{\textit{e}}_P \sim E_T(c_p^i)$, $\textbf{\textit{e}}_Q \sim Q(x)$, $\textbf{\textit{e}}_G \sim G(\textbf{\textit{e}}_Q, T(c_s))$
\State {\textit{\textbf{Stage 1: Pre-training Projection Head}}}
\State Randomly initialize a projection head $P_\theta(\cdot)$ which project $\textbf{\textit{e}}_I \rightarrow \textbf{\textit{e}}_G$
\While{not converged}
    \State Generate embeddings: $\textbf{\textit{e}}_I \leftarrow E_I(x)$, $\textbf{\textit{e}}_G \leftarrow G(Q(\textbf{\textit{e}}_I), T(c_s))$
    \State Compute the prediction based on precise $\textbf{\textit{e}}_G$: $\textbf{\textit{e}}_I' \leftarrow P(\textbf{\textit{e}}_I)$ 
    \State Update $P_\theta$: $P_{\theta} \leftarrow P_{\theta}- \nu \nabla_{P_{\theta}} \mathcal{L}$ where $\mathcal{L} = \|\textbf{\textit{e}}_G - \textbf{\textit{e}}_I'\|^2$
\EndWhile
\State \textbf{return} $P_{\theta}$

\State{\textbf{\textit{Stage 2: Zero-Shot Prediction}}}

\State Trained projection head $P_\theta(\cdot)$
\For{each caption $c_p^i$ in $C_p$}
\State Generate embeddings: $\textbf{\textit{e}}_I = E_I(x)$, $\textbf{\textit{e}}_P = E_T(c_p^i)$
\State Compute cosine similarity: cos\_sim$(P_\theta(\textbf{\textit{e}}_I), P_\theta(\textbf{\textit{e}}_P)) = \frac{P_\theta(\textbf{\textit{e}}_I) \cdot P_\theta(\textbf{\textit{e}}_P)}{\|P_\theta(\textbf{\textit{e}}_I)\|\|P_\theta(\textbf{\textit{e}}_P)\|}$
\EndFor
\State \textbf{return} $c_p^*$ which yields the highest similarity score is the prediction.

\end{algorithmic}
\end{algorithm}

\subsection{Enhanced Zero-shot FER Model}
\label{sec:zeroshotfer}
Our approach leverages the pre-trained CLIP model, which is designed to align images with their corresponding textual descriptions, making it naturally fit for zero-shot tasks. This alignment is accomplished by comparing the features of the images with the classification weights generated by the text encoder. The text encoder utilizes textual descriptions that specify the classes of interest as input. 

Formally, for SFER, given a facial image $x \sim X$ with size $ H \times W $, the CLIP image encoder $E_I(\cdot)$ is used to get the CLIP visual embedding $ \textbf{\textit{e}}_I \in \mathbb{R}^{L} $, where $L$ is the length of the feature vectors. For DFER, given a facial video $x \sim X$, we sample $ T $ frames of size $ H \times W $ to form an input $ \textit{x} \in \mathbb{R}^{T \times 3 \times H \times W} $. For each frame $ x^{t} $, we first utilize the shared CLIP image encoder $E_I(\cdot)$ to extract frame-level feature vectors $ \textbf{\textit{e}}^t \in \mathbb{R}^{L} $, where $ t \in \lbrace 1, 2, \cdots, T \rbrace $. Then average pooling is applied along the temporal axes to get the CLIP visual embedding $ \textbf{\textit{e}}_I \in \mathbb{R}^{L} $. Regarding the text part, we feed the prompt list $C_p^{i}$ into CLIP text encoder $E_T(\cdot)$ to get the CLIP textual embedding $ \textbf{\textit{e}}_P^i \in \mathbb{R}^{L} $, where $ i \in \lbrace 1, 2, \cdots, U \rbrace $ and $U$ denotes the number of the categories. Different from the original CLIP, we adopt a shared projection head $P_\theta(\cdot)$ on both image and text encoders, then the final prediction can be formulated as:
\begin{equation}
\vspace{-0.3em}
\begin{aligned}
p(y = k | x) = \dfrac{\exp(\cos(P_\theta(\textbf{\textit{e}}_I), P_\theta(\textbf{\textit{e}}_P^k))/\tau)}{\sum_{k^{\prime}=1}^{U} \exp(\cos(P_\theta(\textbf{\textit{e}}_I), P_\theta(\textbf{\textit{e}}_P^{k^{\prime}}))/\tau)} 
\end{aligned} 
\label{equ1}
\vspace{-0.3em}
\end{equation}
where $\tau$ is a temperature parameter learned by CLIP and $\cos(\cdot, \cdot)$ denotes cosine similarity.

\subsection{LLM Knowledge Transferring}
\label{sec:knowledgetrans}
The plain CLIP feature space learns general representations, lacking task-specific knowledge \cite{ni2022expanding,zheng2022general}. Therefore, we leverage the task-specific knowledge generated from LLMs to improve the CLIP feature representations. This task-specific knowledge is elicited by an instruction-based strategy. To enhance the CLIP visual feature representation for understanding facial activities, we propose projecting the original CLIP visual features into a domain-aware feature space using a projection head, rather than fine-tuning the CLIP image encoder. The key design principle of this projection head is to maintain the strong vision-language alignment inherent in CLIP while emphasizing crucial task-specific elements and filtering out irrelevant information. To achieve this, we employ a linear layer as the projection head, implemented using a simple MLP layer. The empirical results in \ref{sec:ablation} demonstrate that this single MLP layer outperforms more complex modules, providing an efficient and effective way to improve zero-shot performance.

As shown in the contrastive pre-training phase in Fig. \ref{fig_2}, regarding the left part, given an unlabelled image $x \sim X$ with size $ H \times W $, the frozen CLIP image encoder $E_I(\cdot)$ is first applied to get the CLIP visual embedding $ \textbf{\textit{e}}_I \in \mathbb{R}^{L} $. Then this CLIP visual embedding will go through a learnable projection head $P_\theta(\cdot)$, and the output can be denoted as $P_\theta(\textbf{\textit{e}}_I) \in \mathbb{R}^{L^\prime}$. For the right part, the inputs contain the same unlabelled facial image as the CLIP image encoder and an additional instruction text. The image $x$ will go into the I2T Module denoted as $Q(\cdot)$ including a ViT-L14, a Q-Former, and a projector, and the output embedding can be denoted as $Q(\textbf{\textit{e}}_I)\in \mathbb{R}^{H}$, where $H$ is the vector length of the output. The instruction $c_s$ will be fed into the Text Tokenizer $T(\cdot)$ to obtain the instruction word embedding $T(c_s))\in \mathbb{R}^{W}$, where $W$ is the length of the instruction text. Then the image embedding $Q(\textbf{\textit{e}}_I)$ will be concatenated with instruction text embedding $T(c_s)$ as input tokens for LLM encoder $G(x)$. Then the generated feature embedding $\textbf{\textit{e}}_G \in \mathbb{R}^{L^\prime}$ can be denoted as:
\begin{equation}
\vspace{-0.3em}
\begin{aligned}
\textbf{\textit{e}}_G = G(concat(Q(\textbf{\textit{e}}_I), T(c_s)))
\end{aligned} 
\label{equ2}
\vspace{-0.3em}
\end{equation}

To align the CLIP feature space to the LLM task-specific feature space, we utilize the contrastive loss to optimize the learnable projection head:
\begin{equation}
\vspace{-0.3em}
\begin{aligned}
p(y = k | x) = \dfrac{\exp(\cos(P_\theta(\textbf{\textit{e}}_I^k), \textbf{\textit{e}}_G^k)/\tau)}{\sum_{k^{\prime}=1}^{N} \exp(\cos(P_\theta(\textbf{\textit{e}}_I^{k^{\prime}}), \textbf{\textit{e}}_G^{k^{\prime}})/\tau)} 
\end{aligned} 
\label{equ3}
\vspace{-0.3em}
\end{equation}
where $N$ is the number of the mini-batch.

During the training phase, only the learnable projection head $P_\theta(\cdot)$ which is a matrix with shape $ L \times L^\prime $ will be updated, and the rest of the models are all frozen. In contrast to fine-tuning the entire model end-to-end, this approach has the benefit of retaining the ability of the previous model to extract powerful general-purpose representations but adapt it to the FER domain with very few trainable parameters.

\section{Experiments}
\subsection{Datasets}
We conduct experiments on seven popular in-the-wild facial expression benchmarks, including three static FER datasets: RAF-DB \cite{li2018reliable}, AffectNet \cite{m2017affectnet}, and FERPlus \cite{barsoum2016training}, and four dynamic FER datasets: DFEW \cite{jiang2020dfew}, FERV39k \cite{wang2022ferv39k}, MAFW \cite{liu2022mafw}, and AFEW \cite{dhall2012collecting}.

\textbf{\textit{SFER Datasets:}} \textbf{RAF-DB} \cite{li2018reliable} contains 30,000 facial images annotated with basic or compound expressions. Consistent with most previous work, only images with six basic emotions (\ie, happiness, sadness, anger, surprise, disgust, and fear) and neutral are used, including 3,068 images as test data. \textbf{AffectNet} \cite{m2017affectnet} contains about 450,000 images that are manually annotated with 11 expression categories. The seven expression categories denoted by AffectNet-7 contain six basic emotions and neutral, including 3,500 images as test data. The eight expression categories denoted by AffectNet-8 contain six basic emotions, contempt, neutral. \textbf{FERPlus} \cite{dhall2012collecting} contains 35,887 images and is an extended version of FER2013. Same as AffectNet-8, eight expression categories are used, including 3,589 images as test data.

\textbf{\textit{DFER Datasets:}} \textbf{DFEW} \cite{jiang2020dfew} contains 11,697 in-the-wild video clips, and all the samples have been split into five same-size parts without overlap. Each video is annotated to one of the six basic emotions and neutral. The 5-fold cross-validation is adopted as an evaluation protocol. \textbf{FERV39k} \cite{wang2022ferv39k} contains 38,935 in-the-wild video clips, which is currently the largest in-the-wild DFER dataset. All the video clips were assigned to one of the six basic emotions and neutral. Video clips of all scenes are randomly shuffled and split into training (80\%), and testing (20\%) without overlapping. \textbf{MAFW} \cite{liu2022mafw} contains 10,045 in-the-wild video clips, which is the first large-scale, multi-modal, multi-label affective database with 11 single-expression categories (six basic emotions, contempt, anxiety, helplessness, disappointment, and neutral), 32 multiple-expression categories, and emotional descriptive texts. The 5-fold cross-validation is adopted as an evaluation protocol. \textbf{AFEW} \cite{dhall2012collecting} contains 1,809 in-the-wild video clips, and all the samples have been split into three splits: train (773 video clips), validation (383 video clips), and test (653 video clips). Since the test split is not publicly available, we report results on the validation split. 

\begin{table*}[!t]
\begin{center}
\caption{Zero-shot FER results in comparison with other visual language models. B/32, B/16, L/14 denote ViT-B/32, ViT-B/16, and ViT-L/14, respectively. $^{*}$EmoCLIP was pre-trained on MAFW \cite{liu2022mafw} with facial expression clips and action unit text descriptions without categorical labels. \textbf{Bold} donates the best, \underline{Underline} denotes the second best.}
\vspace{-1em}
\label{tab1}
\scriptsize
\begin{tabular}{@{}c|cccc|cccc@{}}
\toprule
\multirow{2}{*}{Methods} &
  \multicolumn{4}{c|}{Static FER (UAR/WAR)} &
  \multicolumn{4}{c}{Dynamic FER (UAR/WAR)} \\ \cmidrule(l){2-9} 
 &
  \multicolumn{1}{c|}{RAF-DB} &
  \multicolumn{1}{c|}{AffectNet-7} &
  \multicolumn{1}{c|}{AffectNet-8} &
  \multicolumn{1}{c|}{FERPlus} &
  \multicolumn{1}{c|}{DFEW} &
  \multicolumn{1}{c|}{FERV39k} &
  \multicolumn{1}{c|}{MAFW} &
                      AFEW \\ \midrule
BLIP2 (L/14) \cite{li2023blip} &
  \multicolumn{1}{c|}{43.78/44.07} &
  \multicolumn{1}{c|}{32.86/32.87} &
  \multicolumn{1}{c|}{28.53/28.53} &
                      \underline{43.40}/48.14 &
  \multicolumn{1}{c|}{-} &
  \multicolumn{1}{c|}{-} &
  \multicolumn{1}{c|}{-} &
                      - \\
\midrule  
FaRL (B/16) \cite{zheng2022general} &
  \multicolumn{1}{c|}{24.98/38.53} &
  \multicolumn{1}{c|}{26.95/26.95} &
  \multicolumn{1}{c|}{23.50/23.51} &
                      24.27/35.26 &
  \multicolumn{1}{c|}{23.99/25.73} &
  \multicolumn{1}{c|}{20.71/22.87} &
  \multicolumn{1}{c|}{14.98/17.08} &
                      28.17/30.45 \\
FLAVA \cite{singh2022flava} &
  \multicolumn{1}{c|}{14.35/38.69} &
  \multicolumn{1}{c|}{14.26/14.26} &
  \multicolumn{1}{c|}{12.47/12.48} &
                      12.50/28.43 &
  \multicolumn{1}{c|}{14.29/20.91} &
  \multicolumn{1}{c|}{14.28/18.77} &
  \multicolumn{1}{c|}{9.31/3.26} &
                     14.29/16.27 \\ 
VideoCLIP \cite{xu2021videoclip} &
  \multicolumn{1}{c|}{-} &
  \multicolumn{1}{c|}{-} &
  \multicolumn{1}{c|}{-} &
                      - &
  \multicolumn{1}{c|}{18.15/17.52} &
  \multicolumn{1}{c|}{15.35/15.57} &
  \multicolumn{1}{c|}{10.26/11.70} &
                      16.32/17.32 \\
XCLIP (B/32) \cite{ni2022expanding} &
  \multicolumn{1}{c|}{-} &
  \multicolumn{1}{c|}{-} &
  \multicolumn{1}{c|}{-} &
                      - &
  \multicolumn{1}{c|}{20.03/16.06} &
  \multicolumn{1}{c|}{16.80/12.50} &
  \multicolumn{1}{c|}{11.51/13.53} &
                      21.81/22.05 \\
EmotionCLIP \cite{zhang2023learning} &
  \multicolumn{1}{c|}{-} &
  \multicolumn{1}{c|}{-} &
  \multicolumn{1}{c|}{-} &
                      - &
  \multicolumn{1}{c|}{13.77/19.89} &
  \multicolumn{1}{c|}{14.79/19.54} &
  \multicolumn{1}{c|}{9.20/11.65} &
                     14.86/17.06 \\
EmoCLIP$^{*}$ \cite{foteinopoulou2023emoclip} &
  \multicolumn{1}{c|}{-} &
  \multicolumn{1}{c|}{-} &
  \multicolumn{1}{c|}{-} &
                      - &
  \multicolumn{1}{c|}{\underline{36.76/46.27}} &
  \multicolumn{1}{c|}{\underline{26.73}/\textbf{35.30}} &
  \multicolumn{1}{c|}{\textbf{25.86/33.49}} &
                      \underline{36.13/39.90} \\
\midrule
\textbf{Exp-CLIP B/32 (Ours)} &
  \multicolumn{1}{c|}{39.56/42.14} &
  \multicolumn{1}{c|}{31.73/31.73} &
  \multicolumn{1}{c|}{27.63/27.63} &
  \multicolumn{1}{c|}{37.83/37.04} &
  \multicolumn{1}{c|}{24.25/25.87} &
  \multicolumn{1}{c|}{21.48/21.25} &
  \multicolumn{1}{c|}{17.53/20.27} &
                      29.72/31.23 \\
\textbf{Exp-CLIP B/16 (Ours)} &
  \multicolumn{1}{c|}{\underline{48.96/54.50}} &
  \multicolumn{1}{c|}{\underline{39.98/39.98}} &
  \multicolumn{1}{c|}{\underline{34.40/34.40}} &
  \multicolumn{1}{c|}{40.81/\underline{53.02}} &
  \multicolumn{1}{c|}{27.12/31.75} &
  \multicolumn{1}{c|}{24.78/27.99} &
  \multicolumn{1}{c|}{19.17/23.35} &
                      32.84/33.86 \\
\textbf{Exp-CLIP L/14 (Ours)} &
  \multicolumn{1}{c|}{\textbf{58.70/65.37}} &
  \multicolumn{1}{c|}{\textbf{44.27/44.27}} &
  \multicolumn{1}{c|}{\textbf{38.44/38.43}} &
  \multicolumn{1}{c|}{\textbf{48.28/55.42}} &
  \multicolumn{1}{c|}{\textbf{40.16/47.09}} &
  \multicolumn{1}{c|}{\textbf{27.45}/\underline{31.07}} &
  \multicolumn{1}{c|}{\underline{23.95/26.98}} &
                      \textbf{38.72/40.42} \\ \bottomrule
\end{tabular}
\vspace{-1.5em}
\end{center}
\end{table*}
\begin{table*}[!t]
\scriptsize
\vspace{-1.5em}
\begin{center}
\caption{Zero-shot FER results in comparison with CLIP model.}
\vspace{-1.5em}
\label{tab2}
\begin{tabular}{@{}c|cccccccc@{}}
\toprule
\multirow{2}{*}{Methods} &
  \multicolumn{4}{c|}{Static FER (UAR/WAR)} &
  \multicolumn{4}{c}{Dynamic FER (UAR/WAR)} \\ \cmidrule(l){2-9} 
 &
  \multicolumn{1}{c|}{RAF-DB} &
  \multicolumn{1}{c|}{AffectNet-7} &
  \multicolumn{1}{c|}{AffectNet-8} &
  \multicolumn{1}{c|}{FERPlus} &
  \multicolumn{1}{c|}{DFEW} &
  \multicolumn{1}{c|}{FERV39k} &
  \multicolumn{1}{c|}{MAFW} &
                      AFEW \\ \midrule
 &
  \multicolumn{8}{c}{ViT-B/32} \\ \midrule
CLIP \cite{radford2021learning} &
  \multicolumn{1}{c|}{34.97/30.05} &
  \multicolumn{1}{c|}{29.29/29.29} &
  \multicolumn{1}{c|}{25.53/25.53} &
  \multicolumn{1}{c|}{36.05/33.89} &
  \multicolumn{1}{c|}{22.08/21.30} &
  \multicolumn{1}{c|}{18.12/17.38} &
  \multicolumn{1}{c|}{13.39/15.48} &
                      28.59/28.87 \\ 
Exp-CLIP (Ours) &
  \multicolumn{1}{c|}{\textbf{39.56/42.14}} &
  \multicolumn{1}{c|}{\textbf{31.73/31.73}} &
  \multicolumn{1}{c|}{\textbf{27.63/27.63}} &
  \multicolumn{1}{c|}{\textbf{37.83/37.04}} &
  \multicolumn{1}{c|}{\textbf{24.25/25.87}} &
  \multicolumn{1}{c|}{\textbf{21.48/21.25}} &
  \multicolumn{1}{c|}{\textbf{17.53/20.27}} &
                      \textbf{29.72/31.23} \\ \midrule
\textcolor{red}{{$\Uparrow$}} &
  \multicolumn{1}{c|}{\textcolor{red}{4.59/12.09}} &
  \multicolumn{1}{c|}{\textcolor{red}{2.44/2.44}} &
  \multicolumn{1}{c|}{\textcolor{red}{2.10/2.10}} &
  \multicolumn{1}{c|}{\textcolor{red}{1.78/3.15}} &
  \multicolumn{1}{c|}{\textcolor{red}{2.17/4.57}} &
  \multicolumn{1}{c|}{\textcolor{red}{3.36/3.87}} &
  \multicolumn{1}{c|}{\textcolor{red}{4.14/4.79}} &
                      \textcolor{red}{1.13/2.36} \\ \midrule
 &
  \multicolumn{8}{c}{ViT-B/16} \\ \midrule
CLIP \cite{radford2021learning} &
  \multicolumn{1}{c|}{38.66/37.16} &
  \multicolumn{1}{c|}{34.35/34.35} &
  \multicolumn{1}{c|}{29.38/29.38} &
  \multicolumn{1}{c|}{34.33/45.14} &
  \multicolumn{1}{c|}{22.92/28.33} &
  \multicolumn{1}{c|}{19.81/22.51} &
  \multicolumn{1}{c|}{15.45/17.25} &
                      28.21/29.13 \\ 
Exp-CLIP (Ours) &
  \multicolumn{1}{c|}{\textbf{48.96/54.50}} &
  \multicolumn{1}{c|}{\textbf{39.98/39.98}} &
  \multicolumn{1}{c|}{\textbf{34.40/34.40}} &
  \multicolumn{1}{c|}{\textbf{40.81/53.02}} &
  \multicolumn{1}{c|}{\textbf{27.12/31.75}} &
  \multicolumn{1}{c|}{\textbf{24.78/27.99}} &
  \multicolumn{1}{c|}{\textbf{19.17/23.35}} &
                      \textbf{32.84/33.86} \\ \midrule
\textcolor{red}{$\Uparrow$} &
  \multicolumn{1}{c|}{\textcolor{red}{9.11/15.75}} &
  \multicolumn{1}{c|}{\textcolor{red}{5.63/5.63}} &
  \multicolumn{1}{c|}{\textcolor{red}{5.02/5.02}} &
  \multicolumn{1}{c|}{\textcolor{red}{6.48/7.88}}&
  \multicolumn{1}{c|}{\textcolor{red}{4.20/3.42}} &
  \multicolumn{1}{c|}{\textcolor{red}{4.97/5.48}} &
  \multicolumn{1}{c|}{\textcolor{red}{3.72/6.10}} &
                      \textcolor{red}{4.63/4.73} \\ \midrule
 &
  \multicolumn{8}{c}{ViT-L/14} \\ \midrule
CLIP \cite{radford2021learning} &
  \multicolumn{1}{c|}{47.22/41.13} &
  \multicolumn{1}{c|}{34.46/34.47} &
  \multicolumn{1}{c|}{29.96/29.96} &
  \multicolumn{1}{c|}{33.82/46.67} &
  \multicolumn{1}{c|}{32.02/38.12} &
  \multicolumn{1}{c|}{21.97/28.99} &
  \multicolumn{1}{c|}{20.42/25.58} &
                      35.50/38.32 \\
Exp-CLIP (Ours) &
  \multicolumn{1}{c|}{\textbf{58.70/65.37}} &
  \multicolumn{1}{c|}{\textbf{44.27/44.27}} &
  \multicolumn{1}{c|}{\textbf{38.44/38.43}} &
  \multicolumn{1}{c|}{\textbf{48.28/55.42}} &
  \multicolumn{1}{c|}{\textbf{40.16/47.09}} &
  \multicolumn{1}{c|}{\textbf{27.45/31.07}} &
  \multicolumn{1}{c|}{\textbf{23.95/26.98}} &
                      \textbf{38.72/40.42}  \\ \midrule
\textcolor{red}{$\Uparrow$} &
  \multicolumn{1}{c|}{\textcolor{red}{11.48/24.24}} &
  \multicolumn{1}{c|}{\textcolor{red}{9.81/9.80}} &
  \multicolumn{1}{c|}{\textcolor{red}{8.48/8.47}} &
  \multicolumn{1}{c|}{\textcolor{red}{14.46/8.75}} &
  \multicolumn{1}{c|}{\textcolor{red}{8.14/8.97}} &
  \multicolumn{1}{c|}{\textcolor{red}{5.48/2.08}} &
  \multicolumn{1}{c|}{\textcolor{red}{3.53/1.40}} &
                      \textcolor{red}{3.22/2.10} \\ \bottomrule
\end{tabular}
\vspace{-3.5em}
\end{center}
\end{table*}

\textbf{Experiment Settings}: 
In the contrastive learning phase, the training is conducted on CAER-S \cite{lee2019context} training set with 45K facial images. The projection head in our method refers to a matrix with size $512\times4096$ when utilising ViT-B-32/B-16 or size $768\times4096$ when utilising ViT-L-14. The learning rate for the learnable projection head is set as $1\times10^{-3}$. The learnable projection head is trained for 5 epochs in an end-to-end manner. The parameters of the projection head are optimized by the SGD optimizer with a mini-batch size of 512. Our models are trained on the Tesla A100 GPU based on the open-source PyTorch platform. We employ the Flan-T5-XXL \cite{chung2024scaling} with 11B parameters as our LLM. During the zero-shot prediction, for the SFER, all the images are resized to $224 \times 224$ as an input. For the DEER, all the fixed 16 frames of sequence in our experiments follow the sampling strategy in \cite{jiang2020dfew,zhao2021former,li2023intensity}, and then resized to $224 \times 224$ as inputs. To obtain more stable and reliable results, we train models three times with different random seeds and then use the average as the final result.

\textbf{Evaluation Metrics}: Consistent with most previous methods \cite{zhao2021former,li2023intensity,wang2023rethinking,zhang2023weakly}, the weighted average recall (WAR, i.e., accuracy) and the unweighted average recall (UAR, i.e., the accuracy per class divided by the number of classes without considering the number of instances per class) are utilized for evaluating the performance of methods.

\subsection{Comparisons with the State-of-the-Art}
We first compared our method with several large-language models in a zero-shot way, then compared with some supervised SFER and DFER state of the arts. Our results are obtained using the instruction ``\textit{Please play the role of a facial action describer. Objectively describe the detailed facial actions of the person in the image.}'', and the prompt ``a photo of a face with an expression of [class].'' for zero-shot prediction.

BLIP2 \cite{li2023blip} is a state-of-the-art model on various vision-language tasks, to test its zero-shot performance on the FER task, inspired by work \cite{wu2023gpt4vis}, we take the facial images as input and prompt BLIP2 by using: ``\textit{Question: Play the role of a facial expression classification expert and select one category with the highest similarity to the input image. Here are the optional categories: `contempt', `fear', `disgust', `surprise', `anger', `neutral', `sadness', `happiness'. Answer:}''. We then calculate the WAR and UAR based on the output. As shown in Tab. \ref{tab1}, our results outperform the BLIP2 model on all the FER test sets. Specifically, our ViT-B-16-based zero-shot results can beat BLIP2 in most SFER test sets. Furthermore, we also compare our results with several other visual-language models. VideoCLIP \cite{xu2021videoclip}, XCLIP \cite{ni2022expanding}, and EmotionCLIP \cite{zhang2023learning} are proposed for video analysis, video recognition, and video emotion recognition, respectively, in which VideoCLIP was pre-trained on video clips and text transcriptions, XCLIP was pre-trained on action recognition dataset with action labels, and EmotionCLIP was pre-trained on the videos and conversations. These models all learn spatial-temporal features during the training. However, our method with the bare static model can still outperform them on four DFER test sets. Moreover, we also compared our method with CLIP betterment: FaRL and FLAVA, in which the FaRL model was pre-trained on LAION-FACE specializing in the face analysis task and FLAVA improves CLIP zero-shot performance by introducing both unimodal and multimodal architecture. The results in Tab. \ref{tab1} demonstrate our results are superior to these two methods on all of the FER test sets. Compared with the EmoCLIP \cite{foteinopoulou2023emoclip}, which was trained on the MAFW clips and facial action descriptions, our method outperforms it on DFEW, FERV39k, and AFEW test sets in five metrics without explicit text descriptions, indicating the effectiveness of the proposed method.

\subsection{Ablation Analysis}
\label{sec:ablation}
\textbf{Evaluation of the Projection Head Effectiveness:} Our method is built on the CLIP model \cite{radford2021learning}, to evaluate the effectiveness of the learned projection head, we compare our method to the CLIP model from two views: various ViT models and various zero-shot prompts. Our method uses the instruction ``\textit{Please play the role of a facial action describer. Objectively describe the detailed facial actions of the person in the image.}''. As shown in Tab. \ref{tab2}, under different ViT models, our method outperforms the plain CLIP model on all the test sets. With ViT-B-32, our method improves the average UAR by 2.71\% and average WAR by 4.42\% over eight test sets. With ViT-B-16, our method improves the average UAR by 5.47\% and average WAR by 6.75\% over eight test sets. With ViT-L-14, our method improves the average UAR by 8.07\% and average WAR by 8.23\% over eight test sets. The comparison results on varied prompts are shown in the \textit{Appendix Sec. A}.

\begin{table*}[!t]
\begin{center}
\scriptsize
\caption{Zero-shot FER results of our method with different instructions.}
\label{tab3}
\vspace{-1em}
\begin{tabular}{@{}c|cccccccc@{}}
\toprule
\multirow{2}{*}{} &
  \multicolumn{4}{c|}{Static FER (UAR/WAR)} &
  \multicolumn{4}{c}{Dynamic FER (UAR/WAR)} \\ \cmidrule(l){2-9}
 &
  \multicolumn{1}{c|}{RAF-DB} &
  \multicolumn{1}{c|}{AffectNet-7} &
  \multicolumn{1}{c|}{AffectNet-8} &
  \multicolumn{1}{c|}{FERPlus} &
  \multicolumn{1}{c|}{DFEW} &
  \multicolumn{1}{c|}{FERV39k} &
  \multicolumn{1}{c|}{MAFW} &
                      AFEW \\ \midrule
 &
  \multicolumn{8}{c}{ViT-B/32} \\ \midrule
Task-unrelated &
  \multicolumn{1}{c|}{33.70/34.77} &
  \multicolumn{1}{c|}{30.08/30.09} &
  \multicolumn{1}{c|}{26.10/26.11} &
  \multicolumn{1}{c|}{33.03/35.43} &
  \multicolumn{1}{c|}{21.57/20.74} &
  \multicolumn{1}{c|}{19.02/17.69} &
  \multicolumn{1}{c|}{14.71/17.51} &
                      29.42/30.88 \\
Task-related &
  \multicolumn{1}{c|}{\textbf{38.86/43.98}} &
  \multicolumn{1}{c|}{\textbf{31.38/31.38}} &
  \multicolumn{1}{c|}{\textbf{27.75/27.75}} &
  \multicolumn{1}{c|}{\textbf{39.44/37.01}} &
  \multicolumn{1}{c|}{\textbf{24.94/28.52}} &
  \multicolumn{1}{c|}{\textbf{21.84/23.32}} &
  \multicolumn{1}{c|}{\textbf{17.79/20.85}} &
                      \textbf{30.73/32.55} \\ \midrule
\textcolor{red}{$\Uparrow$} &
  \multicolumn{1}{c|}{\textcolor{red}{5.16/9.21}} &
  \multicolumn{1}{c|}{\textcolor{red}{1.30/1.30}} &
  \multicolumn{1}{c|}{\textcolor{red}{1.64/1.64}} &
  \multicolumn{1}{c|}{\textcolor{red}{6.41/1.58}} &
  \multicolumn{1}{c|}{\textcolor{red}{3.37/7.78}} &
  \multicolumn{1}{c|}{\textcolor{red}{2.83/5.62}} &
  \multicolumn{1}{c|}{\textcolor{red}{3.08/3.34}} &
                      \textcolor{red}{1.31/1.66} \\ \midrule    
 &
  \multicolumn{8}{c}{ViT-B/16} \\ \midrule
Task-unrelated &
  \multicolumn{1}{c|}{41.51/40.97} &
  \multicolumn{1}{c|}{36.27/36.28} &
  \multicolumn{1}{c|}{31.29/31.30} &
  \multicolumn{1}{c|}{38.70/48.27} &
  \multicolumn{1}{c|}{24.69/28.20} &
  \multicolumn{1}{c|}{21.67/23.86} &
  \multicolumn{1}{c|}{17.19/19.69} &
                      30.20/31.06 \\
Task-related &
  \multicolumn{1}{c|}{\textbf{48.84/54.68}} &
  \multicolumn{1}{c|}{\textbf{39.46/39.46}} &
  \multicolumn{1}{c|}{\textbf{34.04/34.04}} &
  \multicolumn{1}{c|}{\textbf{40.19/53.82}} &
  \multicolumn{1}{c|}{\textbf{26.40/31.64}} &
  \multicolumn{1}{c|}{\textbf{25.11/28.57}} &
  \multicolumn{1}{c|}{\textbf{18.98/23.64}} &
                      \textbf{31.50/32.63} \\ \midrule
\textcolor{red}{$\Uparrow$} &
  \multicolumn{1}{c|}{\textcolor{red}{7.33/13.71}} &
  \multicolumn{1}{c|}{\textcolor{red}{3.19/3.18}} &
  \multicolumn{1}{c|}{\textcolor{red}{2.74/2.74}} &
  \multicolumn{1}{c|}{\textcolor{red}{1.49/5.55}} &
  \multicolumn{1}{c|}{\textcolor{red}{1.71/3.44}} &
  \multicolumn{1}{c|}{\textcolor{red}{3.44/4.71}} &
  \multicolumn{1}{c|}{\textcolor{red}{1.79/3.96}} &
                      \textcolor{red}{1.30/1.58} \\ \midrule  
 &
  \multicolumn{8}{c}{ViT-L/14} \\ \midrule

Task-unrelated &
  \multicolumn{1}{c|}{57.32/61.26} &
  \multicolumn{1}{c|}{43.14/43.14} &
  \multicolumn{1}{c|}{37.20/37.20} &
  \multicolumn{1}{c|}{46.79/54.16} &
  \multicolumn{1}{c|}{37.94/43.40} &
  \multicolumn{1}{c|}{25.86/30.23} &
  \multicolumn{1}{c|}{23.54/26.47} &
                      38.08/40.33 \\

Task-related &
  \multicolumn{1}{c|}{\textbf{58.08/65.73}} &
  \multicolumn{1}{c|}{\textbf{44.41/44.41}} &
  \multicolumn{1}{c|}{\textbf{38.24/38.24}} &
  \multicolumn{1}{c|}{\textbf{47.70/55.56}} &
  \multicolumn{1}{c|}{\textbf{40.59/47.19}} &
  \multicolumn{1}{c|}{\textbf{27.90/31.51}} &
  \multicolumn{1}{c|}{\textbf{24.16/27.27}} &
                      \textbf{39.81/41.21} \\ \midrule
\textcolor{red}{$\Uparrow$} &
  \multicolumn{1}{c|}{\textcolor{red}{0.76/4.47}} &
  \multicolumn{1}{c|}{\textcolor{red}{1.27/1.27}} &
  \multicolumn{1}{c|}{\textcolor{red}{1.04/1.04}} &
  \multicolumn{1}{c|}{\textcolor{red}{0.90/1.40}} &
  \multicolumn{1}{c|}{\textcolor{red}{2.65/3.79}} &
  \multicolumn{1}{c|}{\textcolor{red}{2.04/1.27}} &
  \multicolumn{1}{c|}{\textcolor{red}{0.62/0.81}} &
                      \textcolor{red}{1.73/0.87} \\ \bottomrule 
\end{tabular}
\end{center}
\vspace{-1.5em}
\end{table*}

\textbf{Evaluation of Different Instructions:} To verify the effectiveness of the instruction adopted in our method, we conducted experiments on two types of instructions: task-related and task-unrelated. Regarding the task-unrelated instructions, we consider both empty and random text as input and take the average of them as the final results. For the task-related instruction, we provide results with three instructions and their average. As shown in Tab. \ref{tab3}, the task-related instructions are superior to task-unrelated instructions on both SFER and DFER, in which there are 3.14\% average UAR improvements and 4.02\% average WAR improvements over eight test sets with ViT-B-32, 2.87\% average UAR improvements and 4.86\% average WAR improvements over eight test sets with ViT-B-16, and 1.38\% average UAR improvements and 1.87\% average WAR improvements over eight test sets with ViT-L-14. We believe that task-related instructions facilitate LLMs to generate semantic features with task-aware ability, providing better objects for projection head optimisation. The detailed results on task-related and task-unrelated instructions are shown in the \textit{Appendix Sec. B}.

\begin{table*}[!t]
\begin{center}
\caption{Zero-shot FER results of our method with different projection heads. 2-MLP and 3-MLP denote 2-layer and 3-layer multilayer perceptron, respectively. TE denotes Transformer encoder. LM denotes the learnable linear matrix.}
\label{tab4}
\vspace{-1em}
\scriptsize
\begin{tabular}{@{}c|cccc|cccc@{}}
\toprule
\multirow{2}{*}{Projection Heads} &
  \multicolumn{4}{c|}{Static FER (UAR/WAR)} &
  \multicolumn{4}{c}{Dynamic FER (UAR/WAR)} \\ \cmidrule(l){2-9} 
 &
  \multicolumn{1}{c|}{RAF-DB} &
  \multicolumn{1}{c|}{AffectNet-7} &
  \multicolumn{1}{c|}{AffectNet-8} &
  FERPlus &
  \multicolumn{1}{c|}{DFEW} &
  \multicolumn{1}{c|}{FERV39k} &
  \multicolumn{1}{c|}{MAFW} &
  AFEW \\ \midrule
2-MLP &
  \multicolumn{1}{c|}{45.05/59.49} &
  \multicolumn{1}{c|}{35.46/35.47} &
  \multicolumn{1}{c|}{29.73/29.73} &
  \multicolumn{1}{c|}{39.66/46.61} &
  \multicolumn{1}{c|}{31.61/37.35} &
  \multicolumn{1}{c|}{25.21/26.91} &
  \multicolumn{1}{c|}{17.62/18.92} &
                      29.92/31.76 \\
3-MLP &
  \multicolumn{1}{c|}{42.28/61.05} &
  \multicolumn{1}{c|}{34.95/34.95} &
  \multicolumn{1}{c|}{30.18/30.18} &
  \multicolumn{1}{c|}{36.60/51.04} &
  \multicolumn{1}{c|}{29.67/36.56} &
  \multicolumn{1}{c|}{25.30/27.25} &
  \multicolumn{1}{c|}{17.81/19.47} &
                      31.91/33.07 \\
TE &
  \multicolumn{1}{c|}{34.22/53.13} &
  \multicolumn{1}{c|}{28.89/28.89} &
  \multicolumn{1}{c|}{23.88/23.88} &
  \multicolumn{1}{c|}{33.39/45.20} &
  \multicolumn{1}{c|}{29.33/37.36} &
  \multicolumn{1}{c|}{26.06/\textbf{32.31}} &
  \multicolumn{1}{c|}{17.94/23.20} &
                      27.16/28.08 \\ \midrule
\textbf{LM} &
  \multicolumn{1}{c|}{\textbf{58.70/65.37}} &
  \multicolumn{1}{c|}{\textbf{44.27/44.27}} &
  \multicolumn{1}{c|}{\textbf{38.44/38.43}} &
  \multicolumn{1}{c|}{\textbf{48.28/55.42}} &
  \multicolumn{1}{c|}{\textbf{40.16/47.09}} &
  \multicolumn{1}{c|}{\textbf{27.45}/31.07} &
  \multicolumn{1}{c|}{\textbf{23.95/26.98}} &
                      \textbf{38.72/40.42} \\ \bottomrule
                  
\end{tabular}
\vspace{-3em}
\end{center}
\end{table*}

\textbf{Evaluation of Different Projection Heads:} In our method, to align the original CLIP feature representations into task-specific LLM knowledge space, we propose to utilize a projection head as a linear mapping to fulfil this. To investigate if the improvement of the complexity of the projectors can further improve the alignment, we also conduct experiments with 2-layer MLP, 3-layer MLP, and Transformer Encoders. As shown in Tab. \ref{tab4}, even with a simply learnable matrix and fewer parameters, our strategy is more powerful and efficient. Compared to the Transformer Encoders, except for the WAR results on FERV39k, ours shows much better results than more complex Transformer Encoders. We believe that more complex non-linear models will entail overfitting to the LLMs feature space, performing worse in downstream zero-shot FER tasks. The linear projection head adopted in our method is more suitable and interpretable for mapping the original CLIP feature space to a task-specific one, maintaining the capacity of the existing model to capture robust, versatile representations while tailoring it to the FER domain with minimal trainable parameters.


\textbf{Evaluation of Varied Image and Text Encoders:} To investigate if the LLM can do better than our method on zero-shot FER tasks, we conducted experiments using the LLM as either an image encoder (with I2T module), text encoder, or both. As shown in Tab. \ref{tab5}, adopting the LLM as either an image encoder or text encoder results in a worse zero-shot preference. Even when using the LLM as both an image and text encoder, the results remain undesirable and significantly lower than the plain CLIP model and our method. We believe the LLM excels in the semantic perception of images but performs poorly in recognition tasks. By contrast, based on the CLIP model, our method is more efficient and accurate in zero-shot FER tasks.

\begin{table*}[!t]
\begin{center}
\caption{Zero-shot FER results of varied image and text encoders. PH denotes the projection head trained in our method.}
\vspace{-1em}
\label{tab5}
\scriptsize
\begin{tabular}{@{}c|c|cccc|cccc@{}}
\toprule
\multirow{2}{*}{\begin{tabular}[c]{@{}c@{}}Image\\ Encoder\end{tabular}} &
  \multirow{2}{*}{\begin{tabular}[c]{@{}c@{}}Text\\ Encoder\end{tabular}} &
  \multicolumn{4}{c|}{Static FER} &
  \multicolumn{4}{c}{Dynamic FER} \\ \cmidrule(l){3-10} 
 &
   &
  \multicolumn{1}{c|}{RAF-DB} &
  \multicolumn{1}{c|}{AffectNet-7} &
  \multicolumn{1}{c|}{AffectNet-8} &
  \multicolumn{1}{c|}{FERPlus} &
  \multicolumn{1}{c|}{DFEW} &
  \multicolumn{1}{c|}{FERV39k} &
  \multicolumn{1}{c|}{MAFW} &
  AFEW \\ \midrule
CLIP &
  CLIP &
  \multicolumn{1}{c|}{47.22/41.13} &
  \multicolumn{1}{c|}{34.46/34.47} &
  \multicolumn{1}{c|}{29.96/29.96} &
  \multicolumn{1}{c|}{33.82/46.67} &
  \multicolumn{1}{c|}{32.02/38.12} &
  \multicolumn{1}{c|}{21.97/28.99} &
  \multicolumn{1}{c|}{20.42/25.58} &
                      35.50/38.32 \\
CLIP+PH &
  LLM &
  \multicolumn{1}{c|}{24.03/38.62} &
  \multicolumn{1}{c|}{24.03/24.04} &
  \multicolumn{1}{c|}{19.03/19.03} &
  \multicolumn{1}{c|}{21.86/37.07} &
  \multicolumn{1}{c|}{21.31/15.31} &
  \multicolumn{1}{c|}{18.02/18.72} &
  \multicolumn{1}{c|}{10.10/5.76} &
  19.33/19.16 \\
LLM &
  CLIP+PH &
  \multicolumn{1}{c|}{25.81/29.11} &
  \multicolumn{1}{c|}{19.36/19.35} &
  \multicolumn{1}{c|}{17.62/17.60} &
  \multicolumn{1}{c|}{22.59/25.34} &
  \multicolumn{1}{c|}{13.93/3.77} &
  \multicolumn{1}{c|}{15.95/10.69} &
  \multicolumn{1}{c|}{9.66/8.90} &
  19.11/17.06 \\
LLM &
  LLM &
  \multicolumn{1}{c|}{19.46/41.23} &
  \multicolumn{1}{c|}{17.00/17.00} &
  \multicolumn{1}{c|}{14.88/14.88} &
  \multicolumn{1}{c|}{18.01/32.23} &
  \multicolumn{1}{c|}{16.97/22.73} &
  \multicolumn{1}{c|}{15.18/19.28} &
  \multicolumn{1}{c|}{10.25/14.66} &
                      17.94/19.42 \\
CLIP+PH &
  CLIP+PH &
  \multicolumn{1}{c|}{\textbf{58.70/65.37}} &
  \multicolumn{1}{c|}{\textbf{44.27/44.27}} &
  \multicolumn{1}{c|}{\textbf{38.44/38.43}} &
  \multicolumn{1}{c|}{\textbf{48.28/55.42}} &
  \multicolumn{1}{c|}{\textbf{40.16/47.09}} &
  \multicolumn{1}{c|}{\textbf{27.45/31.07}} &
  \multicolumn{1}{c|}{\textbf{23.95/26.98}} &
                      \textbf{38.72/40.42}\\\bottomrule
\end{tabular}
\end{center}
\vspace{-2em}
\end{table*}
\begin{figure*}[!t]
	\centering
	\includegraphics[scale=0.53]{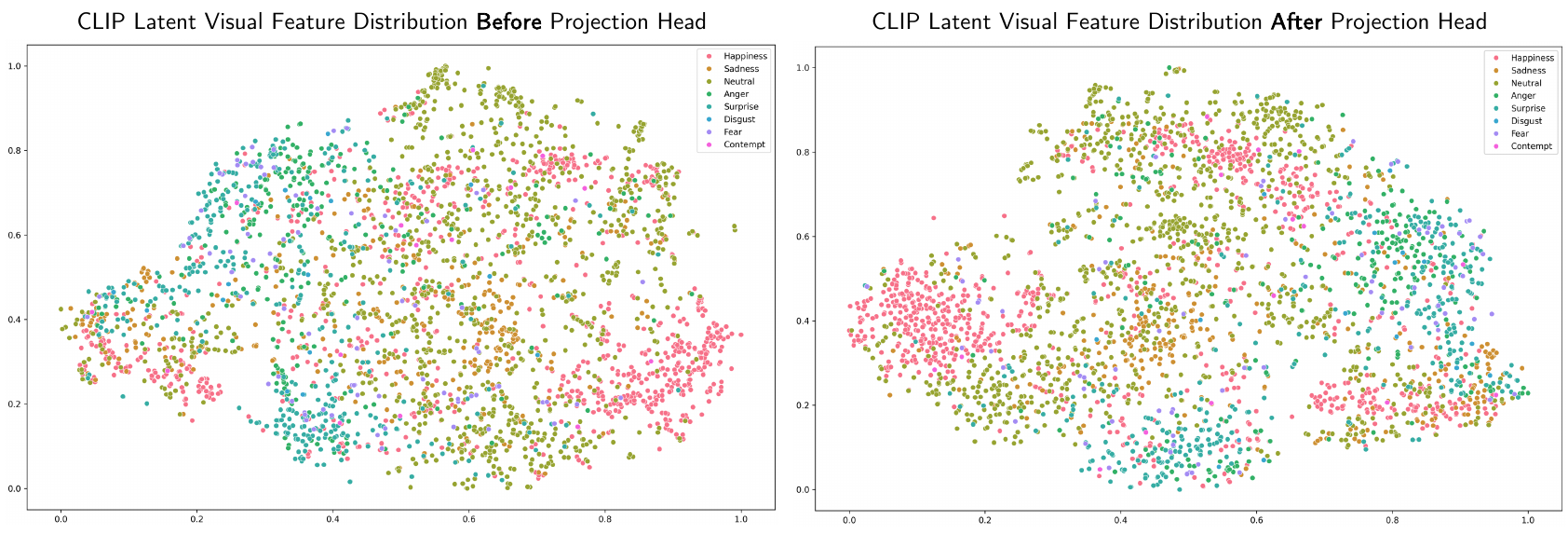}
        \vspace{-1em}
	\caption{CLIP latent visual feature distribution on FERPlus dataset. Both results are based on ViT-L-14.}
	\label{fig_4}
\vspace{-1.5em}
\end{figure*}

\subsection{Visualizations}
To prove the benefit of the enhanced CLIP visual features, we utilize t-SNE \cite{van2008visualizing} to illustrate the feature distribution of the original CLIP feature representations and projected feature representations. Different from the supervised methods learned with annotations which can gather samples of the same category compactly, zero-shot visual feature representations are learned by aligning the cross-modality representations, presenting a weak performance in gathering samples compactly. Despite this, from the visual feature distributions in Fig. \ref{fig_4}, we can see the samples of the same category can be better gathered after the projection, especially for emotions of happiness and disgust. This demonstrates the model learns more discriminative latent visual features after the projection head, and can also explain why our method obtains superior results to the original plain CLIP model. In addition, we calculate the normalized feature variance of each emotion on FERPlus in the \textit{Appendix}. The projected features demonstrate a lower variance than the CLIP feature space, proving the projected features are more concentrated. We also provide the confusion matrix comparison in the \textit{Appendix}, which shows that our method captures more nuanced facial expression features.

\section{Limitations}
The proposed Exp-CLIP improves zero-shot FER results by incorporating domain knowledge from LLMs. However, our method still faces several limitations due to the following assumptions: 1) LLMs can adequately capture and describe the subtle nuances of facial expressions, and 2) LLMs can generate outputs that are well-grounded in the input data, minimizing hallucinations. To address 1), we utilize the Flan-T5-XXL model with 11B parameters, which provides more reliable and detailed knowledge. For 2), we use feature representations extracted after the LLM encoder, bypassing the decoder to mitigate the influence of hallucinations. In addition, our method captures more nuanced facial expression features derived from LLMs, e.g. performing better in distinguishing between neutral and other emotions. However, our zero-shot method still makes some incorrect predictions, particularly when distinguishing between similar emotions like Anger and Disgust, as well as Surprise and Fear, especially on hard or ambiguous samples.
\section{Conclusion}
To improve the zero-shot FER performance of the vision-language model, this paper proposes an Exp-CLIP boosted by transferring the task knowledge from large language models (LLMs). To achieve this knowledge transfer, a projection head is employed to align the projected visual representations with task-specific semantic meanings derived from the LLM encoder, and the text instruction-based strategy is employed to customize the LLM knowledge. As a result, the projected visual representations learn task-aware features, facilitating the downstream zero-shot FER task. The training was conducted with one learnable matrix on unlabelled facial data, which is efficient and labour-saving. Exhaustive experiments demonstrate the effectiveness of the proposed strategy. The comparison to several vision-language models shows the superior performance of the proposed method in the FER task. Although there is an obvious results gap between the proposed zero-shot methods and supervised methods, this paper shows a promising direction for more generalised and general FER models. 

\textbf{\textit{Acknowledgments}}: This research utilised Queen Mary's Apocrita HPC facility, supported by QMUL Research-IT. Zengqun Zhao is funded by Queen Mary Principal's PhD Studentships. We thank Dr. Ziquan Liu, James Oldfield, and Yimeng Gu for their valuable comments.

{\small
\bibliographystyle{ieee_fullname}
\bibliography{reference}}

\appendix
\newpage

\twocolumn[
\begin{centering}
    \Large \textbf{Enhancing Zero-Shot Facial Expression Recognition by LLM Knowledge Transfer (Appendix)} \\
    \vspace{2em}
\end{centering}
]

\renewcommand{\thesubsection}{\Alph{subsection}}
\subsection{Comparison with CLIP on Varied Prompts}

\begin{table*}[!b]
\begin{center}
\renewcommand\thetable{A}
\caption{Zero-shot FER results with different prompts in comparison with CLIP model. Both single prompts and prompt ensembles are investigated. We employ three types for single prompts, labelled as P-1, P-2, and P-3 respectively. For prompt ensembles, we utilize 5-prompt and 10-prompt configurations, labelled as P-4 and P-5.}
\label{tabA}
\footnotesize
\begin{tabular}{@{}c|c|cccccccc@{}}
\toprule
\multirow{2}{*}{Models} &
  \multirow{2}{*}{Prompts} &
  \multicolumn{4}{c|}{Static   FER (UAR/WAR)} &
  \multicolumn{4}{c}{Dynamic   FER  (UAR/WAR)} \\ \cmidrule(l){3-10} 
 &
   &
  \multicolumn{1}{c|}{RAF-DB} &
  \multicolumn{1}{c|}{AffectNet-7} &
  \multicolumn{1}{c|}{AffectNet-8} &
  \multicolumn{1}{c|}{FERPlus} &
  \multicolumn{1}{c|}{DFEW} &
  \multicolumn{1}{c|}{FERV39k} &
  \multicolumn{1}{c|}{MAFW} &
                      AFEW \\ \midrule
 &
   &
  \multicolumn{8}{c}{ViT-B/32} \\ \midrule
\multirow{5}{*}{CLIP} &
  P-1 &
  \multicolumn{1}{c|}{28.62/23.57} &
  \multicolumn{1}{c|}{26.09/26.09} &
  \multicolumn{1}{c|}{21.86/21.86} &
  \multicolumn{1}{c|}{\textbf{30.29}/19.76} &
  \multicolumn{1}{c|}{21.07/\textbf{26.94}} &
  \multicolumn{1}{c|}{19.34/18.82} &
  \multicolumn{1}{c|}{14.99/16.35} &
                      25.74/26.51 \\
 &
  P-2 &
  \multicolumn{1}{c|}{40.04/35.76} &
  \multicolumn{1}{c|}{32.84/32.84} &
  \multicolumn{1}{c|}{27.86/27.86} &
  \multicolumn{1}{c|}{37.40/27.03} &
  \multicolumn{1}{c|}{23.22/19.85} &
  \multicolumn{1}{c|}{20.71/16.99} &
  \multicolumn{1}{c|}{18.26/18.85} &
                      30.97/29.66 \\
 &
  P-3 &
  \multicolumn{1}{c|}{34.97/30.05} &
  \multicolumn{1}{c|}{29.29/29.29} &
  \multicolumn{1}{c|}{25.53/25.53} &
  \multicolumn{1}{c|}{36.05/33.89} &
  \multicolumn{1}{c|}{22.08/21.30} &
  \multicolumn{1}{c|}{18.12/17.38} &
  \multicolumn{1}{c|}{13.39/15.48} &
                      28.59/28.87 \\
 &
  P-4 &
  \multicolumn{1}{c|}{35.58/25.36} &
  \multicolumn{1}{c|}{29.69/29.69} &
  \multicolumn{1}{c|}{25.53/25.53} &
  \multicolumn{1}{c|}{34.30/25.28} &
  \multicolumn{1}{c|}{20.58/18.36} &
  \multicolumn{1}{c|}{18.56/15.83} &
  \multicolumn{1}{c|}{15.59/17.16} &
                      28.87/28.87 \\
 &
  P-5 &
  \multicolumn{1}{c|}{37.73/30.28} &
  \multicolumn{1}{c|}{31.61/31.61} &
  \multicolumn{1}{c|}{27.31/27.31} &
  \multicolumn{1}{c|}{\textbf{39.85}/29.45} &
  \multicolumn{1}{c|}{23.75/22.48} &
  \multicolumn{1}{c|}{20.90/17.84} &
  \multicolumn{1}{c|}{17.51/19.73} &
                      27.26/27.30 \\ \midrule
\multirow{5}{*}{\textbf{Ours}} &
  P-1 &
  \multicolumn{1}{c|}{\textbf{35.88/41.59}} &
  \multicolumn{1}{c|}{\textbf{29.89/29.90}} &
  \multicolumn{1}{c|}{\textbf{25.93/25.93}} &
  \multicolumn{1}{c|}{28.81/\textbf{38.38}} &
  \multicolumn{1}{c|}{\textbf{21.15}/26.73} &
  \multicolumn{1}{c|}{\textbf{20.03/19.76}} &
  \multicolumn{1}{c|}{\textbf{15.77/19.84}} &
                      \textbf{29.60/31.32} \\ 
 &
  P-2 &
  \multicolumn{1}{c|}{\textbf{42.12/49.12}} &
  \multicolumn{1}{c|}{\textbf{33.61/33.61}} &
  \multicolumn{1}{c|}{\textbf{29.31/29.31}} &
  \multicolumn{1}{c|}{\textbf{39.82/45.55}} &
  \multicolumn{1}{c|}{\textbf{26.38/25.90}} &
  \multicolumn{1}{c|}{\textbf{22.53/20.16}} &
  \multicolumn{1}{c|}{\textbf{18.99/22.14}} &
                      \textbf{31.98/31.93} \\
 &
  P-3 &
  \multicolumn{1}{c|}{\textbf{39.56/42.14}} &
  \multicolumn{1}{c|}{\textbf{31.73/31.73}} &
  \multicolumn{1}{c|}{\textbf{27.63/27.63}} &
  \multicolumn{1}{c|}{\textbf{37.83/37.04}} &
  \multicolumn{1}{c|}{\textbf{24.25/25.87}} &
  \multicolumn{1}{c|}{\textbf{21.48/21.25}} &
  \multicolumn{1}{c|}{\textbf{17.53/20.27}} &
                      \textbf{29.72/31.23} \\
 &
  P-4 &
  \multicolumn{1}{c|}{\textbf{41.04/45.47}} &
  \multicolumn{1}{c|}{\textbf{32.56/32.56}} &
  \multicolumn{1}{c|}{\textbf{28.43/28.43}} &
  \multicolumn{1}{c|}{\textbf{36.24/42.93}} &
  \multicolumn{1}{c|}{\textbf{23.57/24.00}} &
  \multicolumn{1}{c|}{\textbf{21.19/19.98}} &
  \multicolumn{1}{c|}{\textbf{18.14/21.74}} &
                      \textbf{33.69/35.35} \\
 &
  P-5 &
  \multicolumn{1}{c|}{\textbf{41.13/47.63}} &
  \multicolumn{1}{c|}{\textbf{33.62/33.63}} &
  \multicolumn{1}{c|}{\textbf{29.29/29.29}} &
  \multicolumn{1}{c|}{38.38/\textbf{43.89}} &
  \multicolumn{1}{c|}{\textbf{24.48/24.85}} &
  \multicolumn{1}{c|}{\textbf{21.98/20.46}} &
  \multicolumn{1}{c|}{\textbf{18.40/22.97}} &
                      \textbf{32.91/34.65} \\ \midrule
 &
   &
  \multicolumn{8}{c}{ViT-B/16} \\ \midrule
\multirow{5}{*}{CLIP} &
  P-1 &
  \multicolumn{1}{c|}{32.78/29.27} &
  \multicolumn{1}{c|}{30.41/30.41} &
  \multicolumn{1}{c|}{26.26/26.26} &
  \multicolumn{1}{c|}{31.98/27.16} &
  \multicolumn{1}{c|}{23.87/30.44} &
  \multicolumn{1}{c|}{19.23/21.13} &
  \multicolumn{1}{c|}{17.56/\textbf{22.61}} &
                      27.85/30.71 \\
 &
  P-2 &
  \multicolumn{1}{c|}{53.37/50.52} &
  \multicolumn{1}{c|}{38.61/38.61} &
  \multicolumn{1}{c|}{34.03/34.03} &
  \multicolumn{1}{c|}{\textbf{45.10}/39.02} &
  \multicolumn{1}{c|}{\textbf{31.48}/26.45} &
  \multicolumn{1}{c|}{24.80/22.26} &
  \multicolumn{1}{c|}{20.58/23.22} &
                      35.44/34.91 \\ 
 &
  P-3 &
  \multicolumn{1}{c|}{39.85/38.75} &
  \multicolumn{1}{c|}{34.35/34.35} &
  \multicolumn{1}{c|}{29.38/29.38} &
  \multicolumn{1}{c|}{34.33/45.14} &
  \multicolumn{1}{c|}{22.92/28.33} &
  \multicolumn{1}{c|}{19.81/22.51} &
  \multicolumn{1}{c|}{15.45/17.25} &
                      28.21/29.13 \\
 &
  P-4 &
  \multicolumn{1}{c|}{42.69/34.71} &
  \multicolumn{1}{c|}{35.83/35.84} &
  \multicolumn{1}{c|}{30.48/30.48} &
  \multicolumn{1}{c|}{39.25/35.64} &
  \multicolumn{1}{c|}{25.71/28.85} &
  \multicolumn{1}{c|}{20.19/20.31} &
  \multicolumn{1}{c|}{17.60/20.52} &
                      32.56/32.81 \\ 
 &
  P-5 &
  \multicolumn{1}{c|}{41.74/37.42} &
  \multicolumn{1}{c|}{34.49/34.50} &
  \multicolumn{1}{c|}{29.70/29.71} &
  \multicolumn{1}{c|}{34.42/36.95} &
  \multicolumn{1}{c|}{27.89/29.38} &
  \multicolumn{1}{c|}{23.44/23.37} &
  \multicolumn{1}{c|}{19.98/22.24} &
                      33.12/34.12 \\ \midrule
\multirow{5}{*}{\textbf{Ours}} &
  P-1 &
  \multicolumn{1}{c|}{\textbf{41.80/48.58}} &
  \multicolumn{1}{c|}{\textbf{35.43/35.44}} &
  \multicolumn{1}{c|}{\textbf{30.52/30.52}} &
  \multicolumn{1}{c|}{\textbf{38.72/45.29}} &
  \multicolumn{1}{c|}{\textbf{24.65/31.10}} &
  \multicolumn{1}{c|}{\textbf{24.42/27.31}} &
  \multicolumn{1}{c|}{\textbf{19.43}/22.37} &
                      \textbf{37.52/39.02} \\
 &
  P-2 &
  \multicolumn{1}{c|}{\textbf{55.88/60.21}} &
  \multicolumn{1}{c|}{\textbf{39.75/39.76}} &
  \multicolumn{1}{c|}{\textbf{34.44/34.44}} &
  \multicolumn{1}{c|}{42.42/\textbf{53.25}} &
  \multicolumn{1}{c|}{30.78/\textbf{31.52}} &
  \multicolumn{1}{c|}{\textbf{26.72/28.26}} &
  \multicolumn{1}{c|}{\textbf{21.74/26.31}} &
                      \textbf{35.98/35.52} \\
 &
  P-3 &
  \multicolumn{1}{c|}{\textbf{48.96/54.50}} &
  \multicolumn{1}{c|}{\textbf{39.98/39.98}} &
  \multicolumn{1}{c|}{\textbf{34.40/34.40}} &
  \multicolumn{1}{c|}{\textbf{40.81/53.02}} &
  \multicolumn{1}{c|}{\textbf{27.12/31.75}} &
  \multicolumn{1}{c|}{\textbf{24.78/27.99}} &
  \multicolumn{1}{c|}{\textbf{19.17/23.35}} &
                      \textbf{32.84/33.86} \\
 &
  P-4 &
  \multicolumn{1}{c|}{\textbf{52.42/57.26}} &
  \multicolumn{1}{c|}{\textbf{41.10/41.10}} &
  \multicolumn{1}{c|}{\textbf{35.05/35.05}} &
  \multicolumn{1}{c|}{\textbf{43.30/52.94}} &
  \multicolumn{1}{c|}{\textbf{28.52/32.55}} &
  \multicolumn{1}{c|}{\textbf{26.19/29.03}} &
  \multicolumn{1}{c|}{\textbf{21.35/26.84}} &
                      \textbf{36.85/37.71} \\
 &
  P-5 &
  \multicolumn{1}{c|}{\textbf{50.82/57.66}} &
  \multicolumn{1}{c|}{\textbf{40.76/40.76}} &
  \multicolumn{1}{c|}{\textbf{34.88/34.88}} &
  \multicolumn{1}{c|}{\textbf{43.05/54.30}} &
  \multicolumn{1}{c|}{\textbf{30.79/34.59}} &
  \multicolumn{1}{c|}{\textbf{27.53/30.05}} &
  \multicolumn{1}{c|}{\textbf{22.17/27.40}} &
                      \textbf{37.02/38.50} \\ \midrule
 &
   &
  \multicolumn{8}{c}{ViT-L/14} \\ \midrule
\multirow{5}{*}{CLIP} &
  P-1 &
  \multicolumn{1}{c|}{48.20/38.14} &
  \multicolumn{1}{c|}{36.24/36.24} &
  \multicolumn{1}{c|}{29.76/29.76} &
  \multicolumn{1}{c|}{46.59/32.99} &
  \multicolumn{1}{c|}{\textbf{27.74}/24.43} &
  \multicolumn{1}{c|}{19.75/15.59} &
  \multicolumn{1}{c|}{19.77/16.79} &
                      29.93/28.87 \\
 &
  P-2 &
  \multicolumn{1}{c|}{49.37/39.34} &
  \multicolumn{1}{c|}{37.35/37.35} &
  \multicolumn{1}{c|}{33.86/33.86} &
  \multicolumn{1}{c|}{42.06/26.49} &
  \multicolumn{1}{c|}{32.68/20.83} &
  \multicolumn{1}{c|}{20.95/14.54} &
  \multicolumn{1}{c|}{20.02/16.46} &
                      33.82/32.28 \\
 &
  P-3 &
  \multicolumn{1}{c|}{47.22/41.13} &
  \multicolumn{1}{c|}{34.46/34.47} &
  \multicolumn{1}{c|}{29.96/29.96} &
  \multicolumn{1}{c|}{33.82/46.67} &
  \multicolumn{1}{c|}{32.02/38.12} &
  \multicolumn{1}{c|}{21.97/28.99} &
  \multicolumn{1}{c|}{20.42/25.58} &
                      35.50/38.32 \\
 &
  P-4 &
  \multicolumn{1}{c|}{49.79/36.34} &
  \multicolumn{1}{c|}{39.41/39.41} &
  \multicolumn{1}{c|}{33.53/33.53} &
  \multicolumn{1}{c|}{43.65/30.22} &
  \multicolumn{1}{c|}{35.39/29.31} &
  \multicolumn{1}{c|}{21.97/18.85} &
  \multicolumn{1}{c|}{21.84/22.86} &
                      31.61/31.50 \\ 
 &
  P-5 &
  \multicolumn{1}{c|}{52.21/41.36} &
  \multicolumn{1}{c|}{39.72/39.73} &
  \multicolumn{1}{c|}{32.98/32.98} &
  \multicolumn{1}{c|}{43.28/31.40} &
  \multicolumn{1}{c|}{35.85/35.40} &
  \multicolumn{1}{c|}{23.54/24.10} &
  \multicolumn{1}{c|}{23.20/26.99} &
                      34.08/34.65 \\ \midrule
\multirow{5}{*}{\textbf{Ours}} &
  P-1 &
  \multicolumn{1}{c|}{\textbf{49.08/53.13}} &
  \multicolumn{1}{c|}{\textbf{38.05/38.06}} &
  \multicolumn{1}{c|}{\textbf{32.93/32.93}} &
  \multicolumn{1}{c|}{\textbf{48.55/49.76}} &
  \multicolumn{1}{c|}{26.76/\textbf{31.65}} &
  \multicolumn{1}{c|}{\textbf{22.31/25.27}} &
  \multicolumn{1}{c|}{\textbf{20.69/23.92}} &
                      \textbf{32.45/33.95} \\
 &
  P-2 &
  \multicolumn{1}{c|}{\textbf{56.17/60.77}} &
  \multicolumn{1}{c|}{\textbf{42.68/42.68}} &
  \multicolumn{1}{c|}{\textbf{37.46/37.46}} &
  \multicolumn{1}{c|}{\textbf{51.54/55.15}} &
  \multicolumn{1}{c|}{\textbf{39.04/38.48}} &
  \multicolumn{1}{c|}{\textbf{26.45/25.10}} &
  \multicolumn{1}{c|}{\textbf{24.81/28.47}} &
                      \textbf{38.50/38.58} \\
 &
  P-3 &
  \multicolumn{1}{c|}{\textbf{58.70/65.37}} &
  \multicolumn{1}{c|}{\textbf{44.27/44.27}} &
  \multicolumn{1}{c|}{\textbf{38.44/38.43}} &
  \multicolumn{1}{c|}{\textbf{48.28/55.42}} &
  \multicolumn{1}{c|}{\textbf{40.16/47.09}} &
  \multicolumn{1}{c|}{\textbf{27.45/31.07}} &
  \multicolumn{1}{c|}{\textbf{23.95/26.98}} &
                      \textbf{38.72/40.42} \\
 &
  P-4 &
  \multicolumn{1}{c|}{\textbf{56.15/60.23}} &
  \multicolumn{1}{c|}{\textbf{43.48/43.48}} &
  \multicolumn{1}{c|}{\textbf{38.03/38.03}} &
  \multicolumn{1}{c|}{\textbf{50.97/53.30}} &
  \multicolumn{1}{c|}{\textbf{38.15/39.75}} &
  \multicolumn{1}{c|}{\textbf{27.14/27.57}} &
  \multicolumn{1}{c|}{\textbf{25.14/27.72}} &
                      \textbf{40.17/41.29} \\
 &
  P-5 &
  \multicolumn{1}{c|}{\textbf{56.52/61.41}} &
  \multicolumn{1}{c|}{\textbf{43.68/43.68}} &
  \multicolumn{1}{c|}{\textbf{38.11/38.11}} &
  \multicolumn{1}{c|}{\textbf{49.93/53.37}} &
  \multicolumn{1}{c|}{\textbf{39.02/43.57}} &
  \multicolumn{1}{c|}{\textbf{28.70/31.70}} &
  \multicolumn{1}{c|}{\textbf{25.94/30.99}} &
                      \textbf{40.31/42.00} \\ \bottomrule
\end{tabular}
\end{center}
\end{table*}

\begin{table*}[!t]
\begin{center}
\renewcommand\thetable{B}
\caption{Zero-shot FER results of our method with different instructions. Both task-related and task-unrelated instructions are investigated. All models use the prompt ``a photo of a face with an expression of [class].'' for zero-shot prediction.}
\label{tabB}
\footnotesize
\begin{tabular}{@{}c|cccccccc@{}}
\toprule
\multirow{2}{*}{} &
  \multicolumn{4}{c|}{Static FER (UAR/WAR)} &
  \multicolumn{4}{c}{Dynamic FER (UAR/WAR)} \\ \cmidrule(l){2-9}
 &
  \multicolumn{1}{c|}{RAF-DB} &
  \multicolumn{1}{c|}{AffectNet-7} &
  \multicolumn{1}{c|}{AffectNet-8} &
  \multicolumn{1}{c|}{FERPlus} &
  \multicolumn{1}{c|}{DFEW} &
  \multicolumn{1}{c|}{FERV39k} &
  \multicolumn{1}{c|}{MAFW} &
  AFEW \\ \midrule
 &
  \multicolumn{8}{c}{ViT-B/32} \\ \midrule
\multirow{3}{*}{\begin{tabular}[c]{@{}c@{}}Task\\ Unrelated\end{tabular}} &
  \multicolumn{1}{c|}{33.53/43.42} &
  \multicolumn{1}{c|}{31.24/31.24} &
  \multicolumn{1}{c|}{27.18/27.18} &
  \multicolumn{1}{c|}{30.43/36.69} &
  \multicolumn{1}{c|}{22.41/23.60} &
  \multicolumn{1}{c|}{20.28/19.59} &
  \multicolumn{1}{c|}{16.10/20.41} &
                      28.28/29.66 \\
 &
  \multicolumn{1}{c|}{30.49/24.19} &
  \multicolumn{1}{c|}{27.77/27.78} &
  \multicolumn{1}{c|}{24.40/24.41} &
  \multicolumn{1}{c|}{30.54/33.44} &
  \multicolumn{1}{c|}{20.03/16.19} &
  \multicolumn{1}{c|}{18.38/15.42} &
  \multicolumn{1}{c|}{13.21/14.83} &
                      29.33/29.66 \\ 
 &
  \multicolumn{1}{c|}{37.09/36.70} &
  \multicolumn{1}{c|}{31.23/31.24} &
  \multicolumn{1}{c|}{26.73/26.73} &
  \multicolumn{1}{c|}{38.12/36.15} &
  \multicolumn{1}{c|}{22.26/22.43} &
  \multicolumn{1}{c|}{18.39/18.07} &
  \multicolumn{1}{c|}{14.82/17.29} &
                      30.65/33.33 \\ \midrule
Mean &
  \multicolumn{1}{c|}{33.70/34.77} &
  \multicolumn{1}{c|}{30.08/30.09} &
  \multicolumn{1}{c|}{26.10/26.11} &
  \multicolumn{1}{c|}{33.03/35.43} &
  \multicolumn{1}{c|}{21.57/20.74} &
  \multicolumn{1}{c|}{19.02/17.69} &
  \multicolumn{1}{c|}{14.71/17.51} &
                      29.42/30.88 \\ \midrule
Variance &
  \multicolumn{1}{c|}{7.28/63.49} &
  \multicolumn{1}{c|}{2.67/2.66} &
  \multicolumn{1}{c|}{1.48/1.47} &
  \multicolumn{1}{c|}{12.96/2.02} &
  \multicolumn{1}{c|}{1.18/10.58} &
  \multicolumn{1}{c|}{0.80/2.97} &
  \multicolumn{1}{c|}{1.40/5.21} &
                      0.94/2.99 \\ \midrule
\multirow{3}{*}{\begin{tabular}[c]{@{}c@{}}Task\\ Related\end{tabular}} &
  \multicolumn{1}{c|}{39.56/42.14} &
  \multicolumn{1}{c|}{31.73/31.73} &
  \multicolumn{1}{c|}{27.63/27.63} &
  \multicolumn{1}{c|}{37.83/37.04} &
  \multicolumn{1}{c|}{24.25/25.87} &
  \multicolumn{1}{c|}{21.48/21.25} &
  \multicolumn{1}{c|}{17.53/20.27} &
                      29.72/31.23 \\
 &
  \multicolumn{1}{c|}{38.63/45.11} &
  \multicolumn{1}{c|}{31.06/31.07} &
  \multicolumn{1}{c|}{27.58/27.58} &
  \multicolumn{1}{c|}{39.46/37.17} &
  \multicolumn{1}{c|}{25.18/29.48} &
  \multicolumn{1}{c|}{21.94/24.09} &
  \multicolumn{1}{c|}{17.96/21.08} &
                      30.87/32.81 \\
 &
  \multicolumn{1}{c|}{38.39/44.69} &
  \multicolumn{1}{c|}{31.35/31.35} &
  \multicolumn{1}{c|}{28.03/28.03} &
  \multicolumn{1}{c|}{41.03/36.82} &
  \multicolumn{1}{c|}{25.39/30.22} &
  \multicolumn{1}{c|}{22.11/24.61} &
  \multicolumn{1}{c|}{17.89/21.21} &
                      31.61/33.60 \\ \midrule
Mean &
  \multicolumn{1}{c|}{38.86/43.98} &
  \multicolumn{1}{c|}{31.38/31.38} &
  \multicolumn{1}{c|}{27.75/27.75} &
  \multicolumn{1}{c|}{39.44/37.01} &
  \multicolumn{1}{c|}{24.94/28.52} &
  \multicolumn{1}{c|}{21.84/23.32} &
  \multicolumn{1}{c|}{17.79/20.85} &
                      30.73/32.55 \\ \midrule
Variance &
  \multicolumn{1}{c|}{0.26/1.73} &
  \multicolumn{1}{c|}{0.08/0.07} &
  \multicolumn{1}{c|}{0.04/0.04} &
  \multicolumn{1}{c|}{1.70/0.02} &
  \multicolumn{1}{c|}{0.25/3.61} &
  \multicolumn{1}{c|}{0.07/2.18} &
  \multicolumn{1}{c|}{0.04/0.17} &
                      0.61/0.97 \\ \midrule
$\Uparrow$ (mean) &
  \multicolumn{1}{c|}{\textbf{5.16/9.21}} &
  \multicolumn{1}{c|}{\textbf{1.30/1.30}} &
  \multicolumn{1}{c|}{\textbf{1.64/1.64}} &
  \multicolumn{1}{c|}{\textbf{6.41/1.58}} &
  \multicolumn{1}{c|}{\textbf{3.37/7.78}} &
  \multicolumn{1}{c|}{\textbf{2.83/5.62}} &
  \multicolumn{1}{c|}{\textbf{3.08/3.34}} &
                      \textbf{1.31/1.66} \\ \midrule    
 &
  \multicolumn{8}{c}{ViT-B/16} \\ \midrule
\multirow{3}{*}{\begin{tabular}[c]{@{}c@{}}Task\\ Unrelated\end{tabular}} &
  \multicolumn{1}{c|}{41.68/41.53} &
  \multicolumn{1}{c|}{35.01/35.01} &
  \multicolumn{1}{c|}{30.55/30.56} &
  \multicolumn{1}{c|}{39.36/51.16} &
  \multicolumn{1}{c|}{23.48/27.22} &
  \multicolumn{1}{c|}{22.07/24.20} &
  \multicolumn{1}{c|}{16.30/19.67} &
                      30.65/31.23 \\ 
 &
  \multicolumn{1}{c|}{37.23/34.75} &
  \multicolumn{1}{c|}{34.57/34.58} &
  \multicolumn{1}{c|}{29.73/29.73} &
  \multicolumn{1}{c|}{36.25/43.13} &
  \multicolumn{1}{c|}{24.25/28.85} &
  \multicolumn{1}{c|}{20.61/23.44} &
  \multicolumn{1}{c|}{17.13/19.68} &
                      29.09/30.18 \\ 
 &
  \multicolumn{1}{c|}{45.61/46.64} &
  \multicolumn{1}{c|}{39.23/39.24} &
  \multicolumn{1}{c|}{33.60/33.61} &
  \multicolumn{1}{c|}{40.49/50.53} &
  \multicolumn{1}{c|}{26.35/28.54} &
  \multicolumn{1}{c|}{22.34/23.95} &
  \multicolumn{1}{c|}{18.15/19.71} &
                      30.87/31.76 \\ \midrule
Mean &
  \multicolumn{1}{c|}{41.51/40.97} &
  \multicolumn{1}{c|}{36.27/36.28} &
  \multicolumn{1}{c|}{31.29/31.30} &
  \multicolumn{1}{c|}{38.70/48.27} &
  \multicolumn{1}{c|}{24.69/28.20} &
  \multicolumn{1}{c|}{21.67/23.86} &
  \multicolumn{1}{c|}{17.19/19.69} &
                      30.20/31.06 \\ \midrule
Variance &
  \multicolumn{1}{c|}{11.72/23.72} &
  \multicolumn{1}{c|}{4.41/4.42} &
  \multicolumn{1}{c|}{2.77/2.78} &
  \multicolumn{1}{c|}{3.21/13.29} &
  \multicolumn{1}{c|}{1.47/0.50} &
  \multicolumn{1}{c|}{0.58/0.10} &
  \multicolumn{1}{c|}{0.57/0.00} &
                      0.63/0.43 \\ \midrule
\multirow{3}{*}{\begin{tabular}[c]{@{}c@{}}Task\\ Related\end{tabular}} &
  \multicolumn{1}{c|}{48.96/54.50} &
  \multicolumn{1}{c|}{39.98/39.98} &
  \multicolumn{1}{c|}{34.40/34.40} &
  \multicolumn{1}{c|}{40.81/53.02} &
  \multicolumn{1}{c|}{27.12/31.75} &
  \multicolumn{1}{c|}{24.78/27.99} &
  \multicolumn{1}{c|}{19.17/23.35} &
                      32.84/33.86 \\
 &
  \multicolumn{1}{c|}{48.61/54.14} &
  \multicolumn{1}{c|}{39.12/39.13} &
  \multicolumn{1}{c|}{33.63/33.63} &
  \multicolumn{1}{c|}{39.85/53.94} &
  \multicolumn{1}{c|}{25.82/31.21} &
  \multicolumn{1}{c|}{25.28/28.69} &
  \multicolumn{1}{c|}{18.88/23.62} &
                      30.95/32.02 \\
 &
  \multicolumn{1}{c|}{48.94/55.41} &
  \multicolumn{1}{c|}{39.27/39.27} &
  \multicolumn{1}{c|}{34.08/34.08} &
  \multicolumn{1}{c|}{39.91/54.51} &
  \multicolumn{1}{c|}{26.27/31.97} &
  \multicolumn{1}{c|}{25.28/29.03} &
  \multicolumn{1}{c|}{18.90/23.96} &
                      30.71/32.02 \\ \midrule
Mean &
  \multicolumn{1}{c|}{48.84/54.68} &
  \multicolumn{1}{c|}{39.46/39.46} &
  \multicolumn{1}{c|}{34.04/34.04} &
  \multicolumn{1}{c|}{40.19/53.82} &
  \multicolumn{1}{c|}{26.40/31.64} &
  \multicolumn{1}{c|}{25.11/28.57} &
  \multicolumn{1}{c|}{18.98/23.64} &
                      31.50/32.63 \\ \midrule
Variance &
  \multicolumn{1}{c|}{0.03/0.29} &
  \multicolumn{1}{c|}{0.14/0.14} &
  \multicolumn{1}{c|}{0.10/0.10} &
  \multicolumn{1}{c|}{0.19/0.38} &
  \multicolumn{1}{c|}{0.29/0.10} &
  \multicolumn{1}{c|}{0.06/0.19} &
  \multicolumn{1}{c|}{0.02/0.06} &
                      0.91/0.75 \\ \midrule
$\Uparrow$ (mean) &
  \multicolumn{1}{c|}{\textbf{7.33/13.71}} &
  \multicolumn{1}{c|}{\textbf{3.19/3.18}} &
  \multicolumn{1}{c|}{\textbf{2.74/2.74}} &
  \multicolumn{1}{c|}{\textbf{1.49/5.55}} &
  \multicolumn{1}{c|}{\textbf{1.71/3.44}} &
  \multicolumn{1}{c|}{\textbf{3.44/4.71}} &
  \multicolumn{1}{c|}{\textbf{1.79/3.96}} &
                      \textbf{1.30/1.58} \\ \midrule  
 &
  \multicolumn{8}{c}{ViT-L/14} \\ \midrule
\multirow{3}{*}{\begin{tabular}[c]{@{}c@{}}Task\\ Unrelated\end{tabular}} &
  \multicolumn{1}{c|}{55.83/62.32} &
  \multicolumn{1}{c|}{41.41/41.41} &
  \multicolumn{1}{c|}{35.76/35.76} &
  \multicolumn{1}{c|}{44.86/54.72} &
  \multicolumn{1}{c|}{36.39/43.82} &
  \multicolumn{1}{c|}{26.29/31.92} &
  \multicolumn{1}{c|}{23.94/26.94} &
                      37.89/40.16 \\
 &
  \multicolumn{1}{c|}{58.02/62.39} &
  \multicolumn{1}{c|}{44.78/44.78} &
  \multicolumn{1}{c|}{38.36/38.36} &
  \multicolumn{1}{c|}{48.17/54.42} &
  \multicolumn{1}{c|}{38.80/44.34} &
  \multicolumn{1}{c|}{26.24/31.11} &
  \multicolumn{1}{c|}{23.34/27.02} &
                      37.92/40.94 \\
 &
  \multicolumn{1}{c|}{58.12/59.06} &
  \multicolumn{1}{c|}{43.24/43.24} &
  \multicolumn{1}{c|}{37.49/37.48} &
  \multicolumn{1}{c|}{47.35/53.33} &
  \multicolumn{1}{c|}{38.63/42.03} &
  \multicolumn{1}{c|}{25.05/27.67} &
  \multicolumn{1}{c|}{23.35/25.44} &
                      38.44/39.90 \\ \midrule
Mean &
  \multicolumn{1}{c|}{57.32/61.26} &
  \multicolumn{1}{c|}{43.14/43.14} &
  \multicolumn{1}{c|}{37.20/37.20} &
  \multicolumn{1}{c|}{46.79/54.16} &
  \multicolumn{1}{c|}{37.94/43.40} &
  \multicolumn{1}{c|}{25.86/30.23} &
  \multicolumn{1}{c|}{23.54/26.47} &
                      38.08/40.33 \\ \midrule
Variance &
  \multicolumn{1}{c|}{1.12/2.41} &
  \multicolumn{1}{c|}{1.90/1.90} &
  \multicolumn{1}{c|}{1.17/1.17} &
  \multicolumn{1}{c|}{1.98/0.36} &
  \multicolumn{1}{c|}{1.21/0.98} &
  \multicolumn{1}{c|}{0.33/3.39} &
  \multicolumn{1}{c|}{0.08/0.53} &
                      0.06/0.20 \\ \midrule
\multirow{3}{*}{\begin{tabular}[c]{@{}c@{}}Task\\ Related\end{tabular}} &
  \multicolumn{1}{c|}{58.70/65.37} &
  \multicolumn{1}{c|}{44.27/44.27} &
  \multicolumn{1}{c|}{38.44/38.43} &
  \multicolumn{1}{c|}{48.28/55.42} &
  \multicolumn{1}{c|}{40.16/47.09} &
  \multicolumn{1}{c|}{27.45/31.07} &
  \multicolumn{1}{c|}{23.95/26.98} &
                      38.72/40.42 \\
 &
  \multicolumn{1}{c|}{58.35/66.07} &
  \multicolumn{1}{c|}{44.52/44.53} &
  \multicolumn{1}{c|}{38.18/38.18} &
  \multicolumn{1}{c|}{46.93/55.75} &
  \multicolumn{1}{c|}{40.73/47.19} &
  \multicolumn{1}{c|}{28.03/31.62} &
  \multicolumn{1}{c|}{24.27/27.47} &
                      40.00/41.21 \\
 &
  \multicolumn{1}{c|}{57.19/65.74} &
  \multicolumn{1}{c|}{44.44/44.44} &
  \multicolumn{1}{c|}{38.11/38.11} &
  \multicolumn{1}{c|}{47.88/55.50} &
  \multicolumn{1}{c|}{40.88/47.28} &
  \multicolumn{1}{c|}{28.23/31.83} &
  \multicolumn{1}{c|}{24.26/27.37} &
                      40.72/41.99 \\ \midrule
Mean &
  \multicolumn{1}{c|}{58.08/65.73} &
  \multicolumn{1}{c|}{44.41/44.41} &
  \multicolumn{1}{c|}{38.24/38.24} &
  \multicolumn{1}{c|}{47.70/55.56} &
  \multicolumn{1}{c|}{40.59/47.19} &
  \multicolumn{1}{c|}{27.90/31.51} &
  \multicolumn{1}{c|}{24.16/27.27} &
                      39.81/41.21 \\ \midrule
Variance &
  \multicolumn{1}{c|}{0.42/0.08} &
  \multicolumn{1}{c|}{0.01/0.01} &
  \multicolumn{1}{c|}{0.02/0.02} &
  \multicolumn{1}{c|}{0.32/0.02} &
  \multicolumn{1}{c|}{0.10/0.01} &
  \multicolumn{1}{c|}{0.11/0.10} &
  \multicolumn{1}{c|}{0.02/0.04} &
                      0.69/0.41 \\ \midrule
$\Uparrow$ (mean) &
  \multicolumn{1}{c|}{\textbf{0.76/4.47}} &
  \multicolumn{1}{c|}{\textbf{1.27/1.27}} &
  \multicolumn{1}{c|}{\textbf{1.04/1.04}} &
  \multicolumn{1}{c|}{\textbf{0.90/1.40}} &
  \multicolumn{1}{c|}{\textbf{2.65/3.79}} &
  \multicolumn{1}{c|}{\textbf{2.04/1.27}} &
  \multicolumn{1}{c|}{\textbf{0.62/0.81}} &
                      \textbf{1.73/0.87} \\ \bottomrule 
\end{tabular}
\end{center}
\end{table*}

We compare our method with CLIP using different prompts, in which both single prompt and prompt ensembles are considered. Specifically, for a single prompt, we utilize ``\{class name\}.", ``an expression of \{class name\}.", and ``a photo of a face with an expression of \{class name\}.", respectively. For prompt ensembles, we utilize 5-prompt and 10-prompt configurations. All the prompt templates are outlined below:

\begin{itemize}
\small
\itemsep -0.1cm
\item ``\{class name\}.",
\item ``an expression of \{class name\}.",
\item ``a photo of a face exuding \{class name\}.",
\item ``a photo radiating \{class name\} in a person.",
\item ``a photo of a person embodying \{class name\}.",
\item ``a good photo capturing someone's \{class name\}.",
\item ``a photo showing someone immersed in \{class name\}.",
\item ``a photo capturing \{class name\} within an individual.",
\item ``a clean photo showcasing a person's \{class name\}.",
\item ``a photo of a face with an expression of \{class name\}.",
\end{itemize}

For each prompt, we compare our method with the CLIP model one by one. As shown in Tab. \ref{tabA}, except for seven values slightly lower than the CLIP model, our method outperforms the CLIP model on all the rest FER test sets, demonstrating the effectiveness of the proposed method. It should be noted that, compared with the CLIP model, only one extra projection matrix is added on top of the two encoders in our method and this projection matrix was trained in an unsupervised manner.

\begin{figure*}[!t]
\centering
\renewcommand\thefigure{A}
    \subfloat[\small CLIP on RAF-DB]{\includegraphics[scale=0.24]{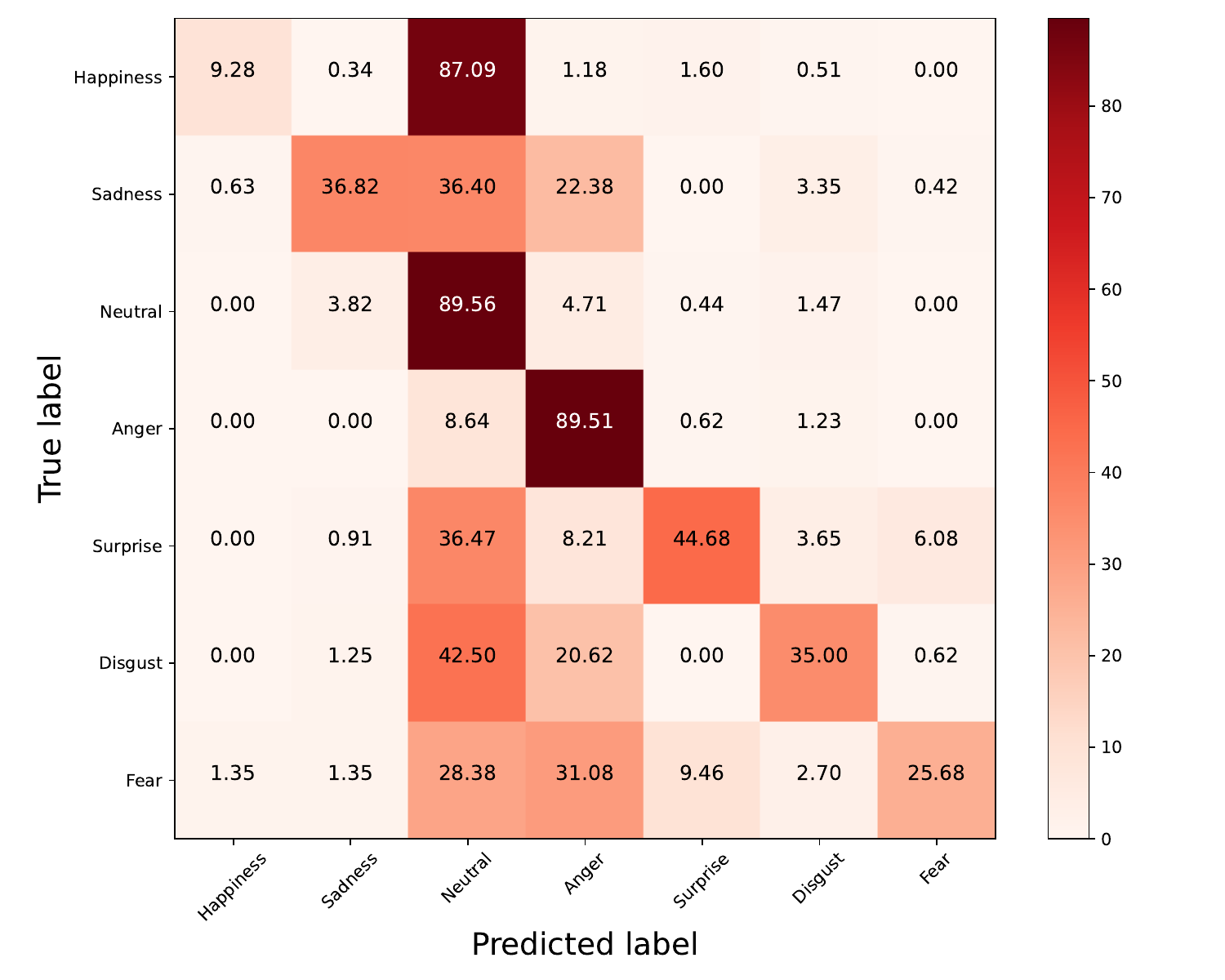}} \hspace{-7.5mm}
    \subfloat[\small CLIP on AffectNet-7]{\includegraphics[scale=0.24]{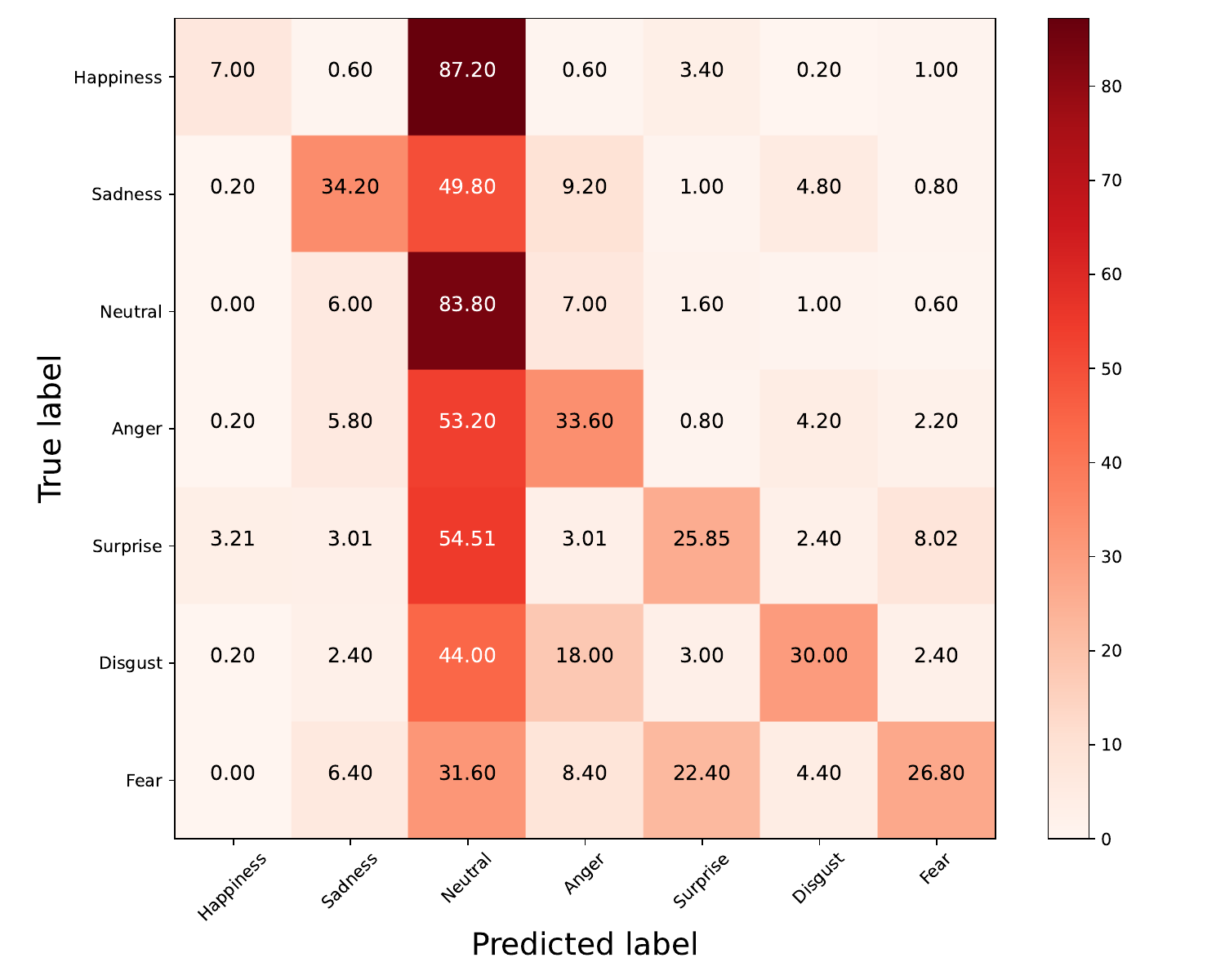}} \hspace{-7mm}
    \subfloat[\small CLIP on FERPlus]{\includegraphics[scale=0.24]{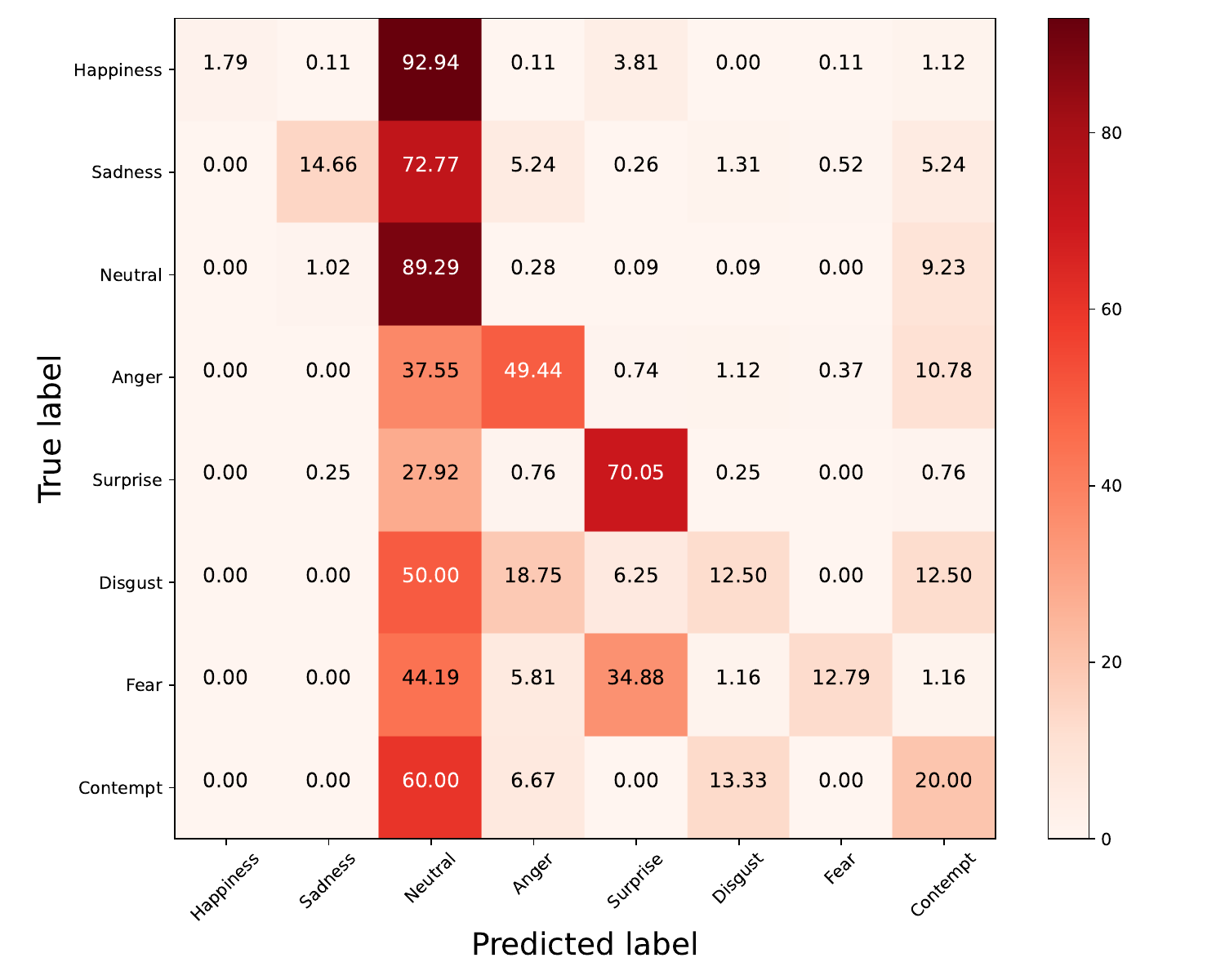}} \hspace{-7mm} 
    \subfloat[\small Ours on RAF-DB]{\includegraphics[scale=0.24]{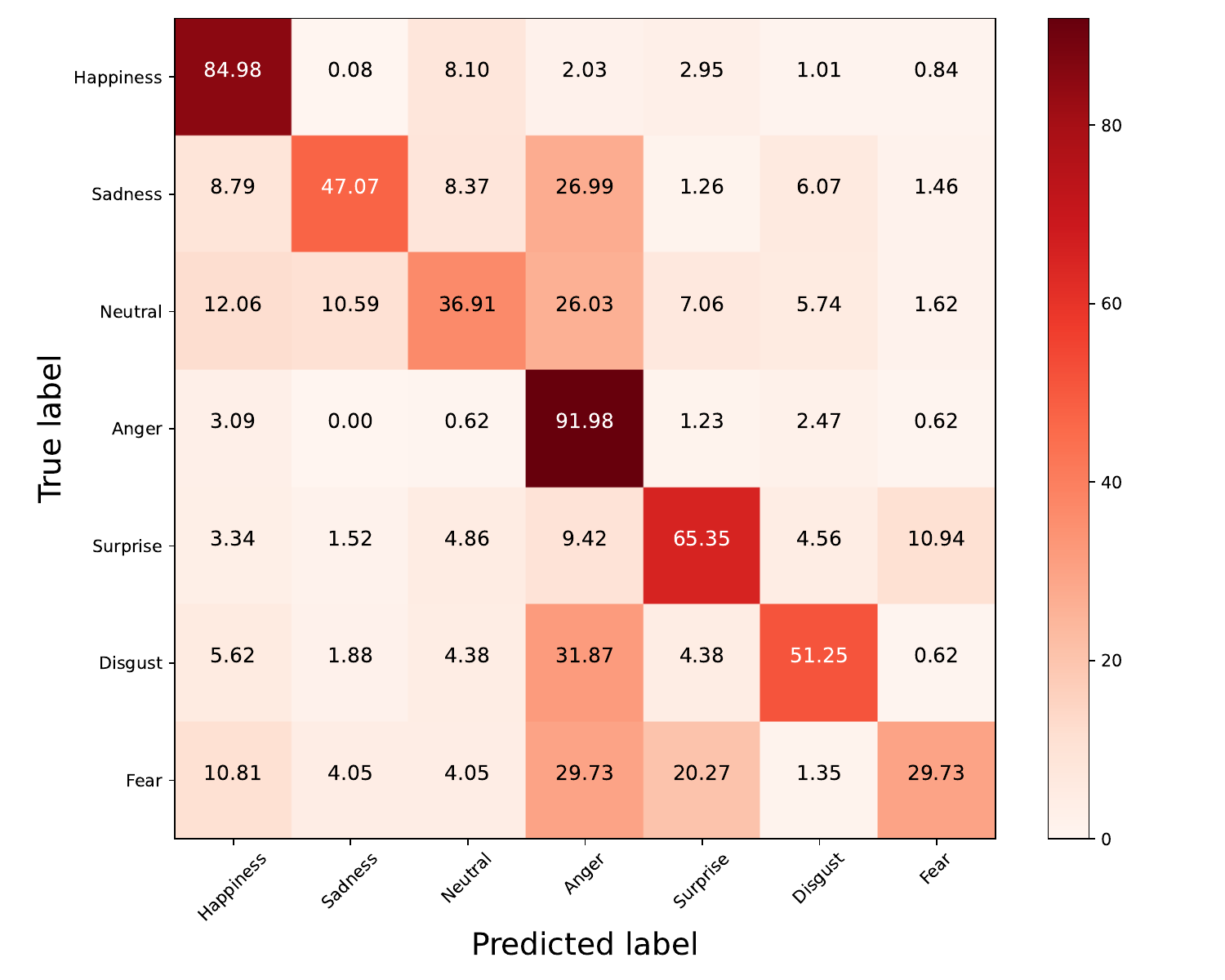}} \hspace{-7mm}
    \subfloat[\small Ours on AffectNet-7]{\includegraphics[scale=0.24]{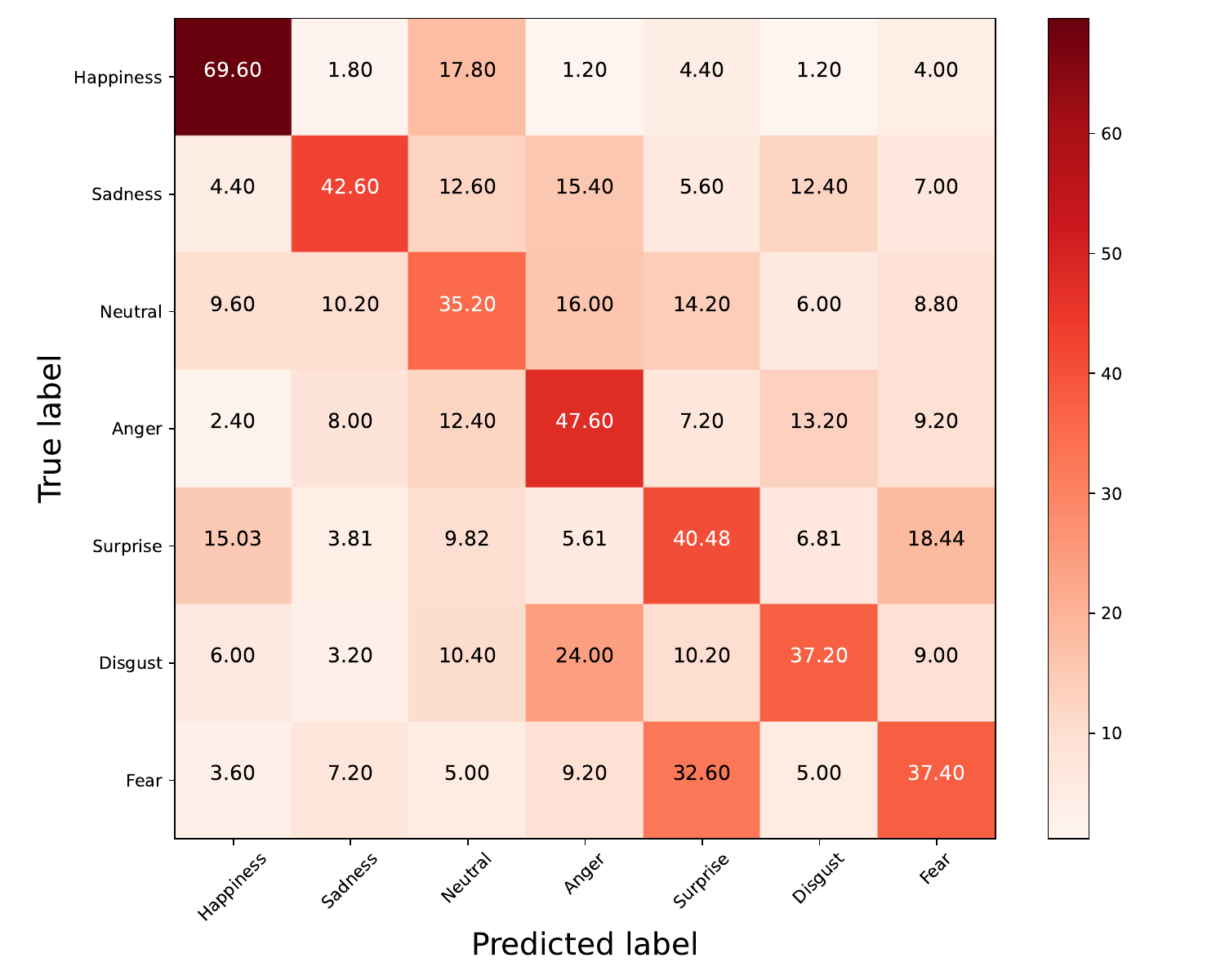}} \hspace{-7mm}
    \subfloat[\small Ours on FERPlus]{\includegraphics[scale=0.24]{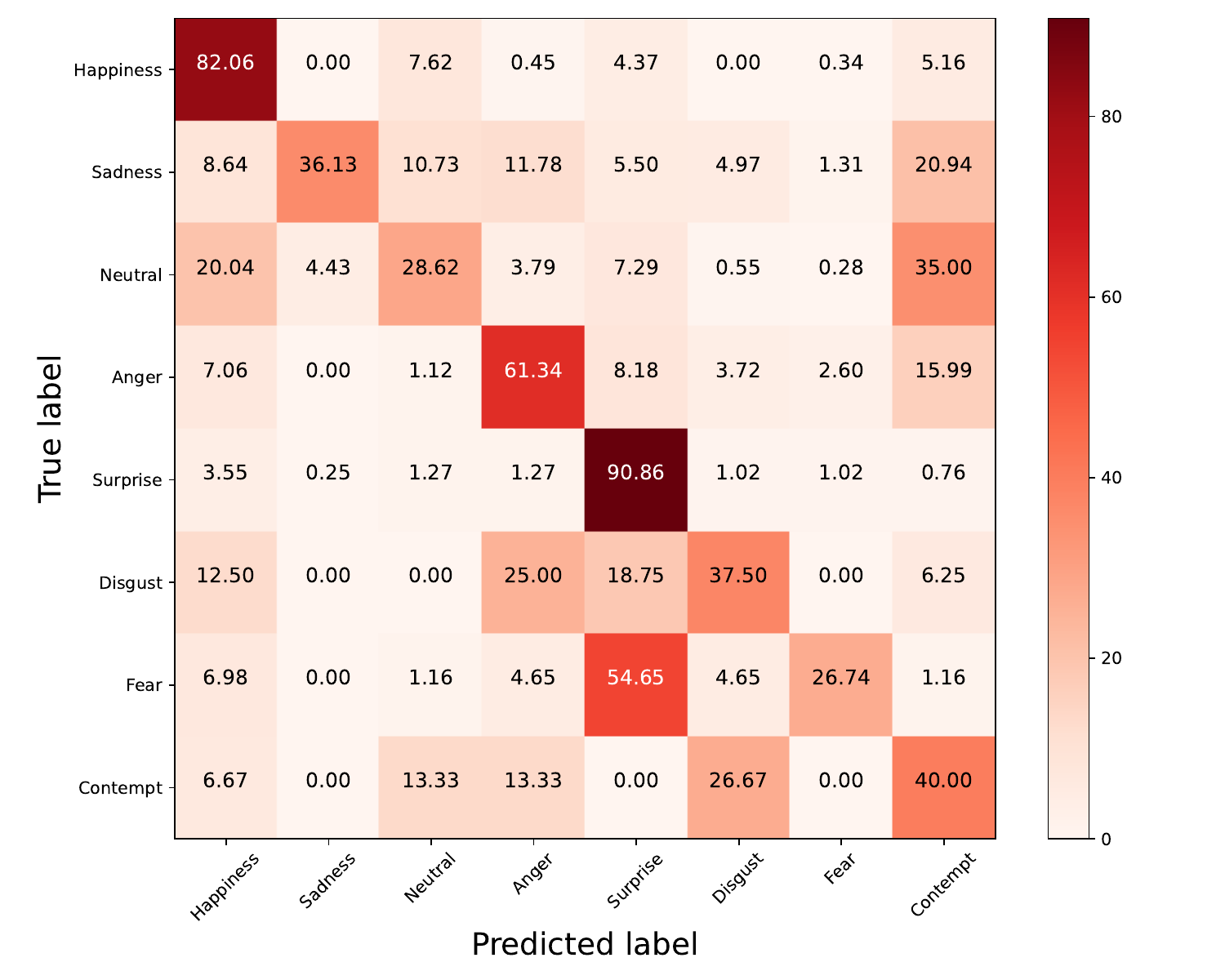}}
    \caption{Confusion matrix compared with CLIP model on static FER datasets.}
    \label{fig_A}
\end{figure*}

\subsection{Evaluation of Different Instructions}
To verify the effectiveness of the instruction adopted in our method, we conducted experiments on two types of instructions: task-related and task-unrelated. Regarding the task-unrelated instructions, we consider both empty and random text as input and take the average of them as the final results. For the task-related instruction, we provide results with three instructions and their average, including ``\textit{Please play the role of a facial action describer. Objectively describe the detailed facial actions of the person in the image.}'', ``\textit{Please play the role of a facial expression recognition expert. Describe the facial expression of the person in the image.}'', and ``\textit{Please describe the detailed facial actions of the person in the image.}''. 

As shown in Tab. \ref{tabB}, the task-related instructions are superior to task-unrelated instructions on both SFER and DFER, in which there are 3.14\% average UAR improvements and 4.02\% average WAR improvements over eight test sets with ViT-B-32, 2.87\% average UAR improvements and 4.86\% average WAR improvements over eight test sets with ViT-B-16, and 1.38\% average UAR improvements and 1.87\% average WAR improvements over eight test sets with ViT-L-14. In addition, the three task-related instructions show a small variance in downstream zero-shot prediction. We believe that task-related instructions facilitate LLMs to generate semantic features with task-aware ability, providing better objects for projection head optimisation.

\begin{figure*}[!t]
\centering
\renewcommand\thefigure{B}
	\subfloat[\small CLIP on DFEW]{\includegraphics[scale=0.24]{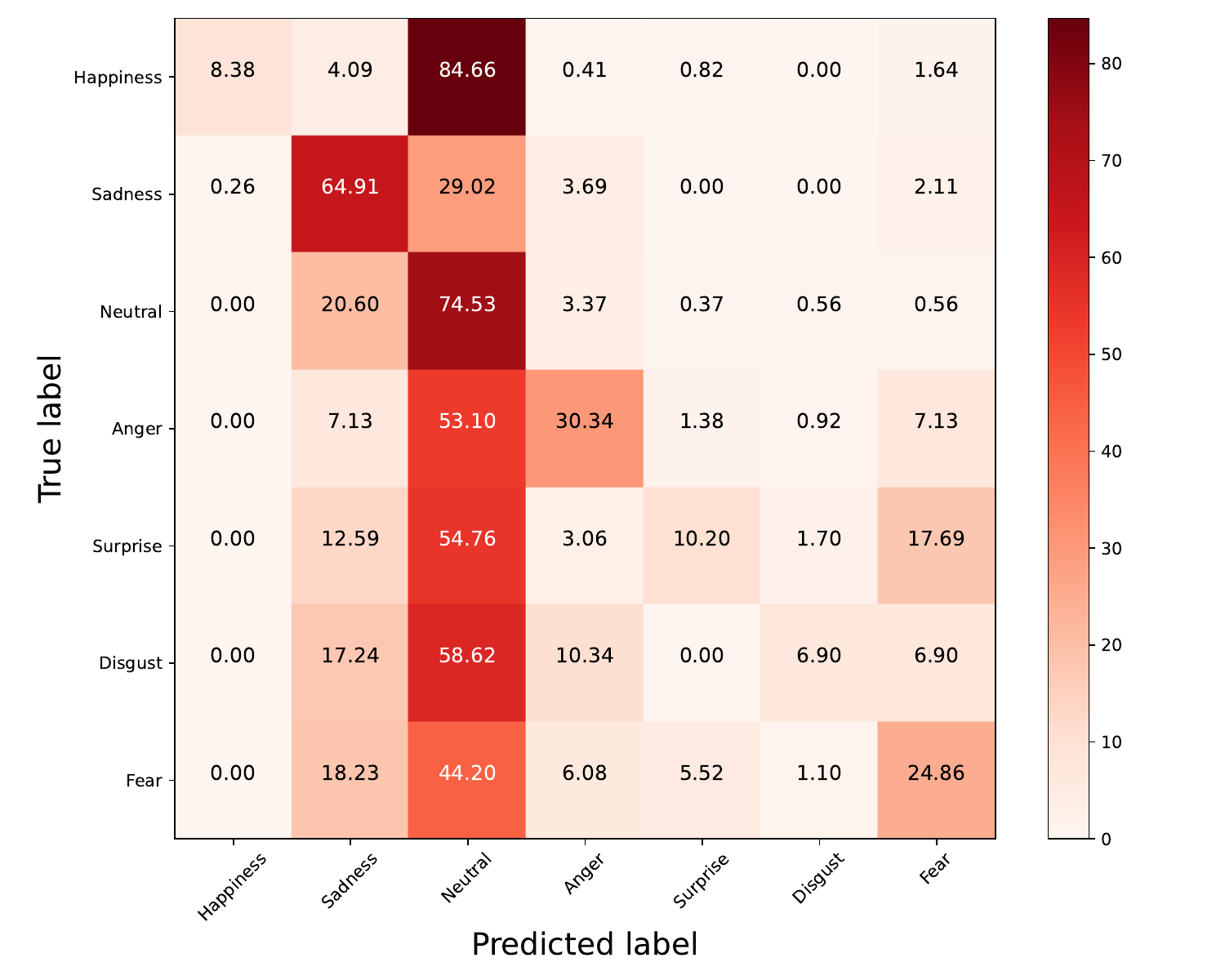}} \hspace{-7mm}
	\subfloat[\small CLIP on FERV39k]{\includegraphics[scale=0.24]{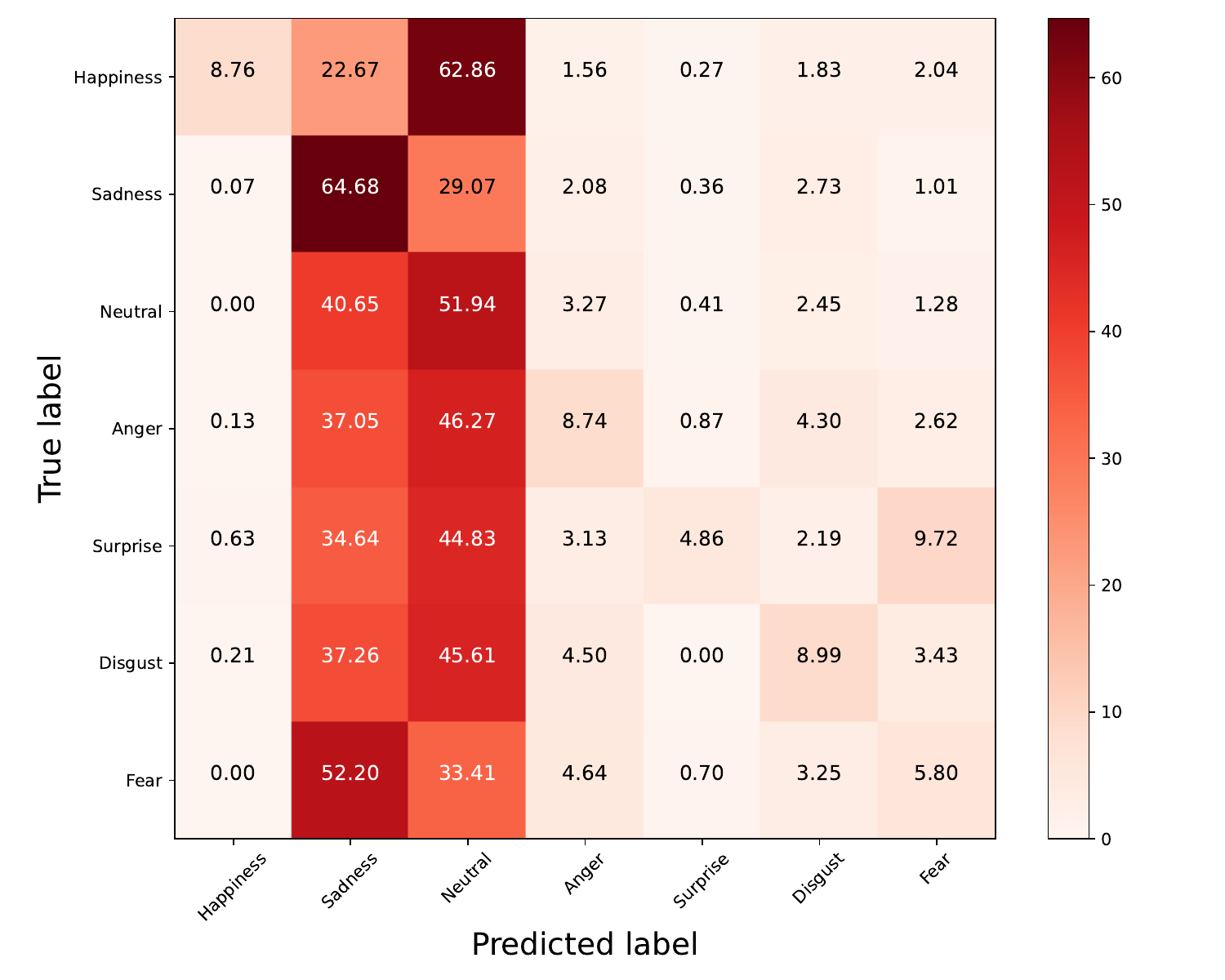}} \hspace{-7mm}
	\subfloat[\small CLIP on MAFW]{\includegraphics[scale=0.24]{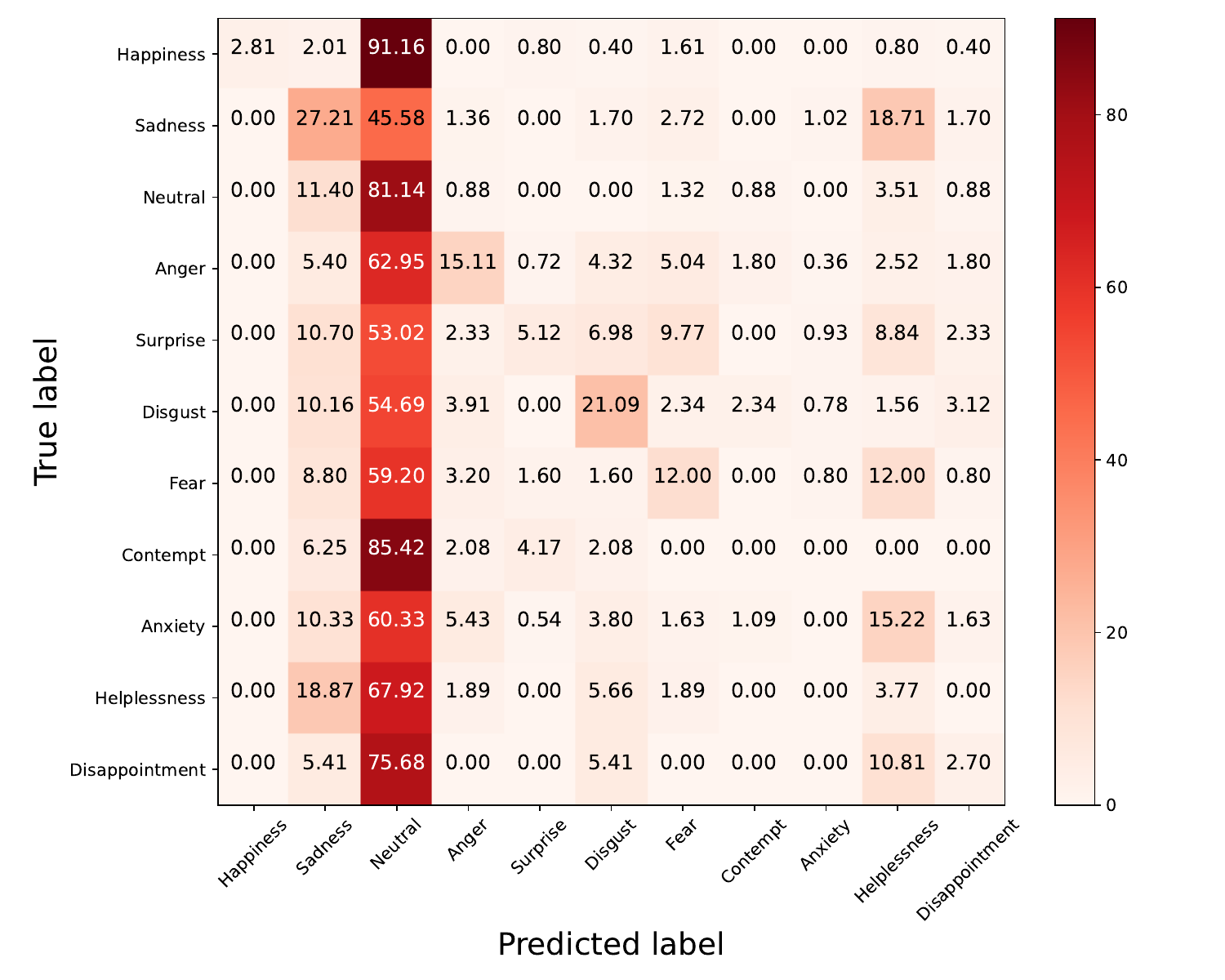}} \hspace{-7mm} 
 	\subfloat[\small Ours on DFEW]{\includegraphics[scale=0.24]{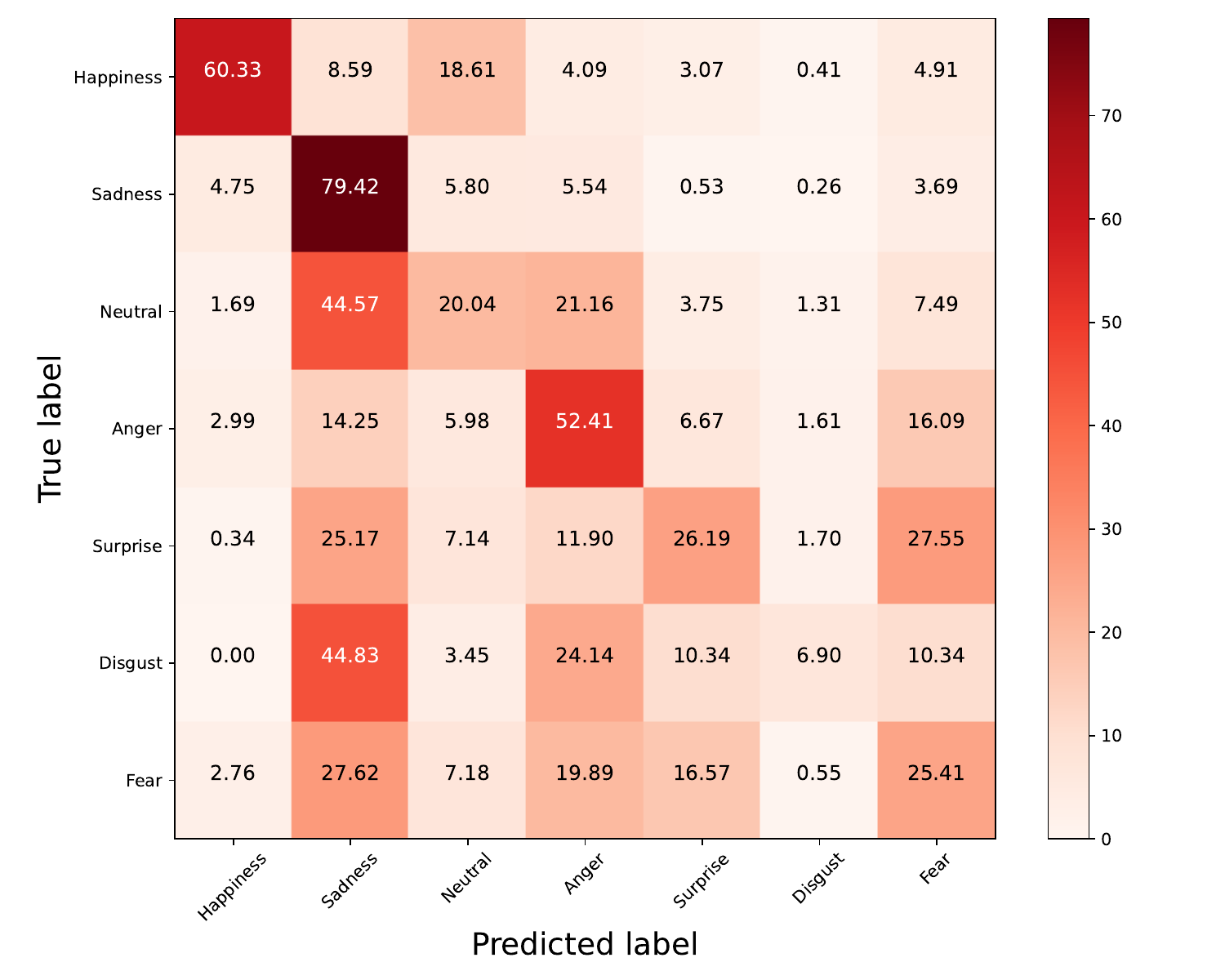}} \hspace{-7mm}
	\subfloat[\small Ours on FERV39k]{\includegraphics[scale=0.24]{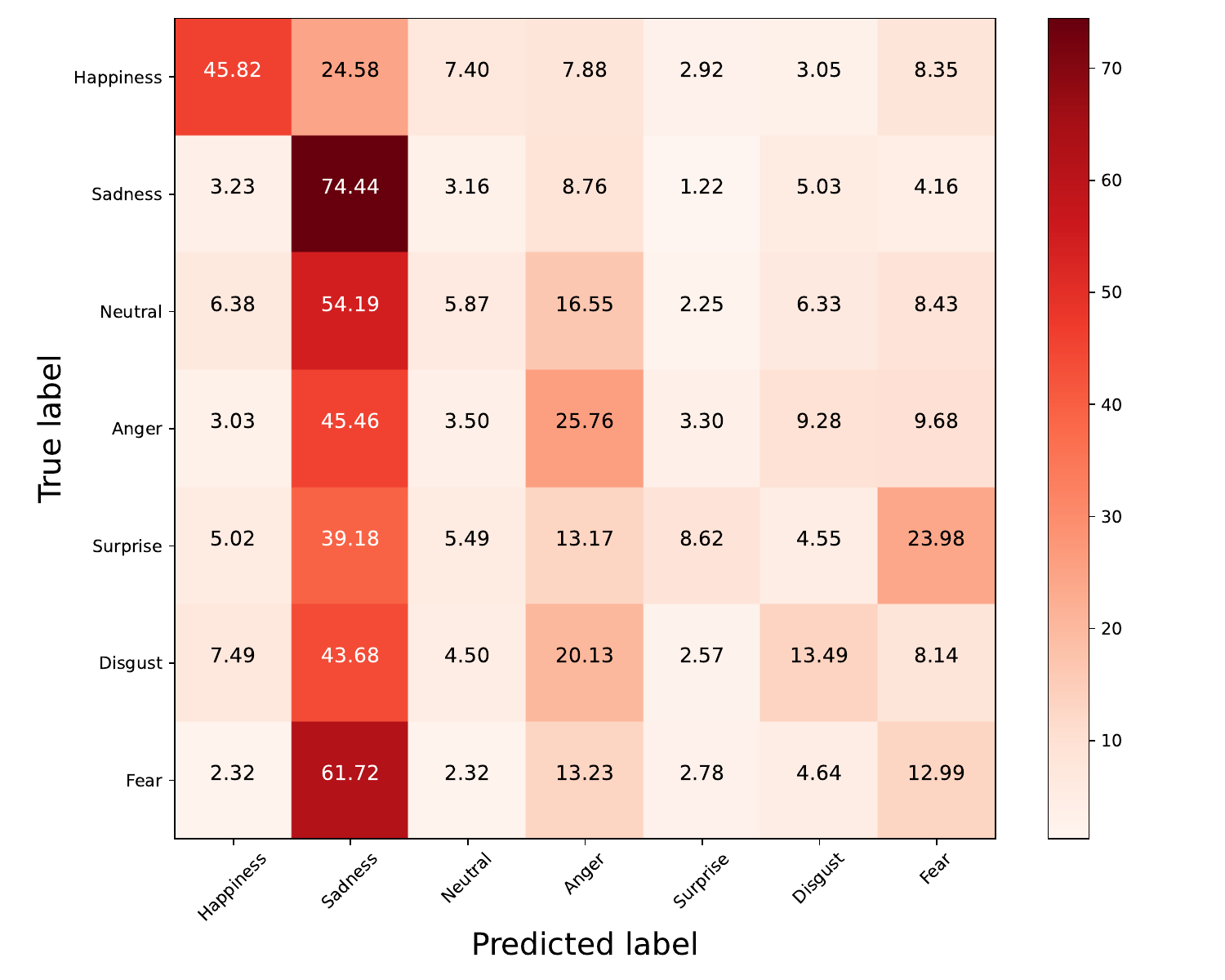}} \hspace{-7mm}
	\subfloat[\small Ours on MAFW]{\includegraphics[scale=0.24]{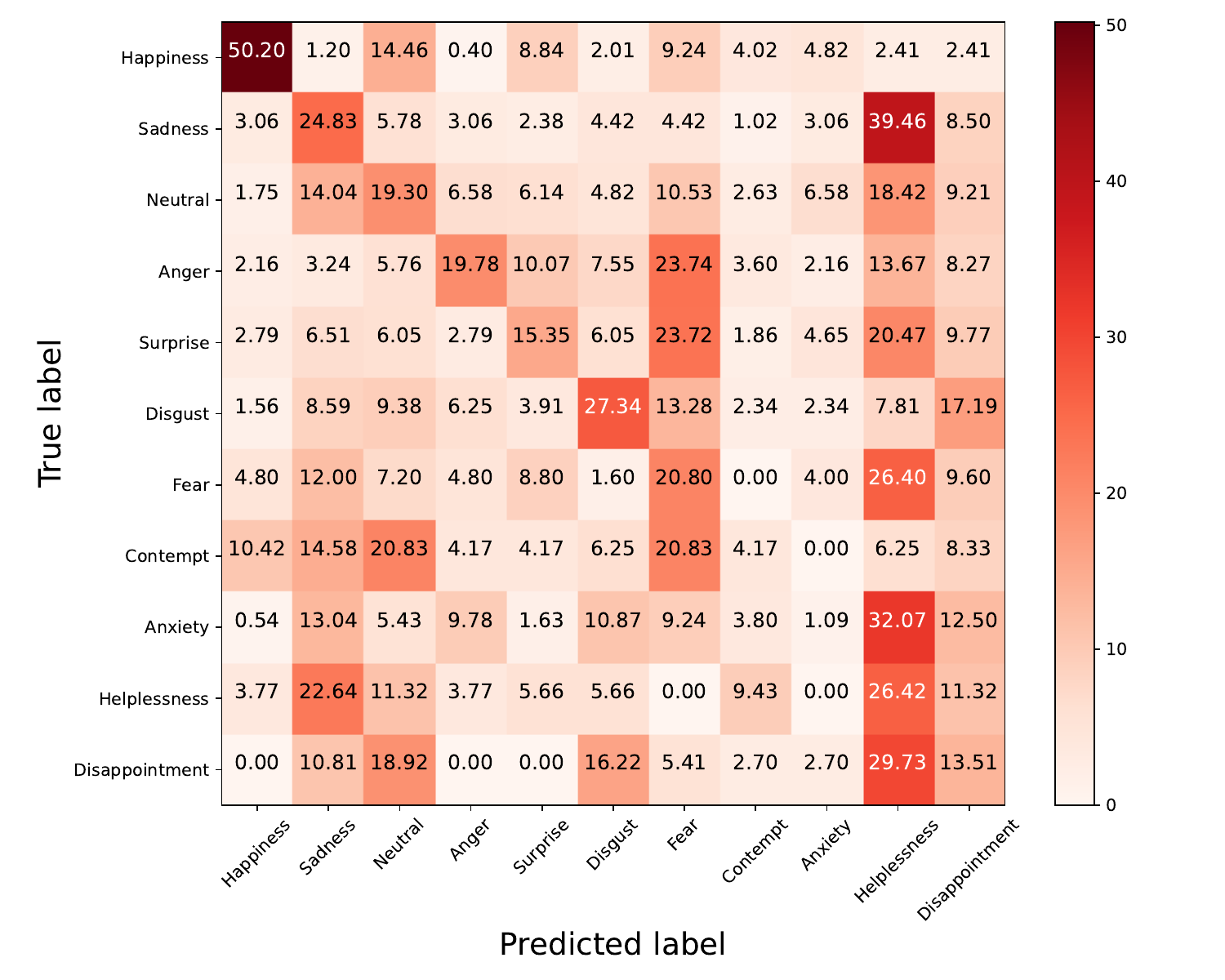}}
	\caption{\small Confusion matrix compared with CLIP model on dynamic FER datasets.}
	\label{fig_B}
\end{figure*}

\subsection{Confusion Matrix Comparison with CLIP}
To offer a more extensive comparison with the CLIP model, we include the learned confusion matrices of both our method and the CLIP model, displayed in Fig. \ref{fig_A} and Fig. \ref{fig_B}. From the confusion matrix on both static FER and dynamic FER datasets, we can see that our method learned better representations for each expression category, resulting in a more balanced accuracy across all categories. Specifically, our method can better distinguish between neutral expressions and other expressions. This suggests that our method captures more nuanced facial expression features derived from LLMs.

\begin{figure}[!t]
\renewcommand\thefigure{C}
	\centering
	\includegraphics[scale=0.4]{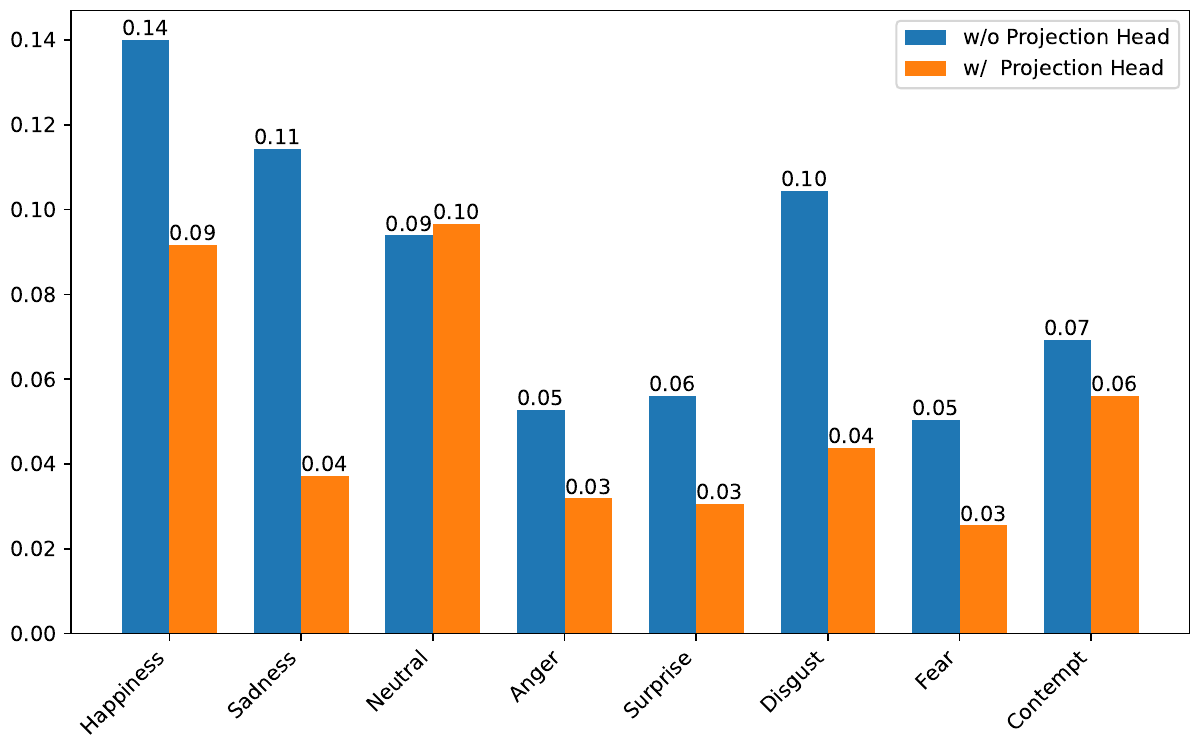}
	\caption{Normalized CLIP latent visual feature variance of each emotion on FERPlus.}
	\label{fig_C}
\end{figure}

\begin{figure}[!t]
\renewcommand\thefigure{D}
	\centering
	\includegraphics[scale=0.24]{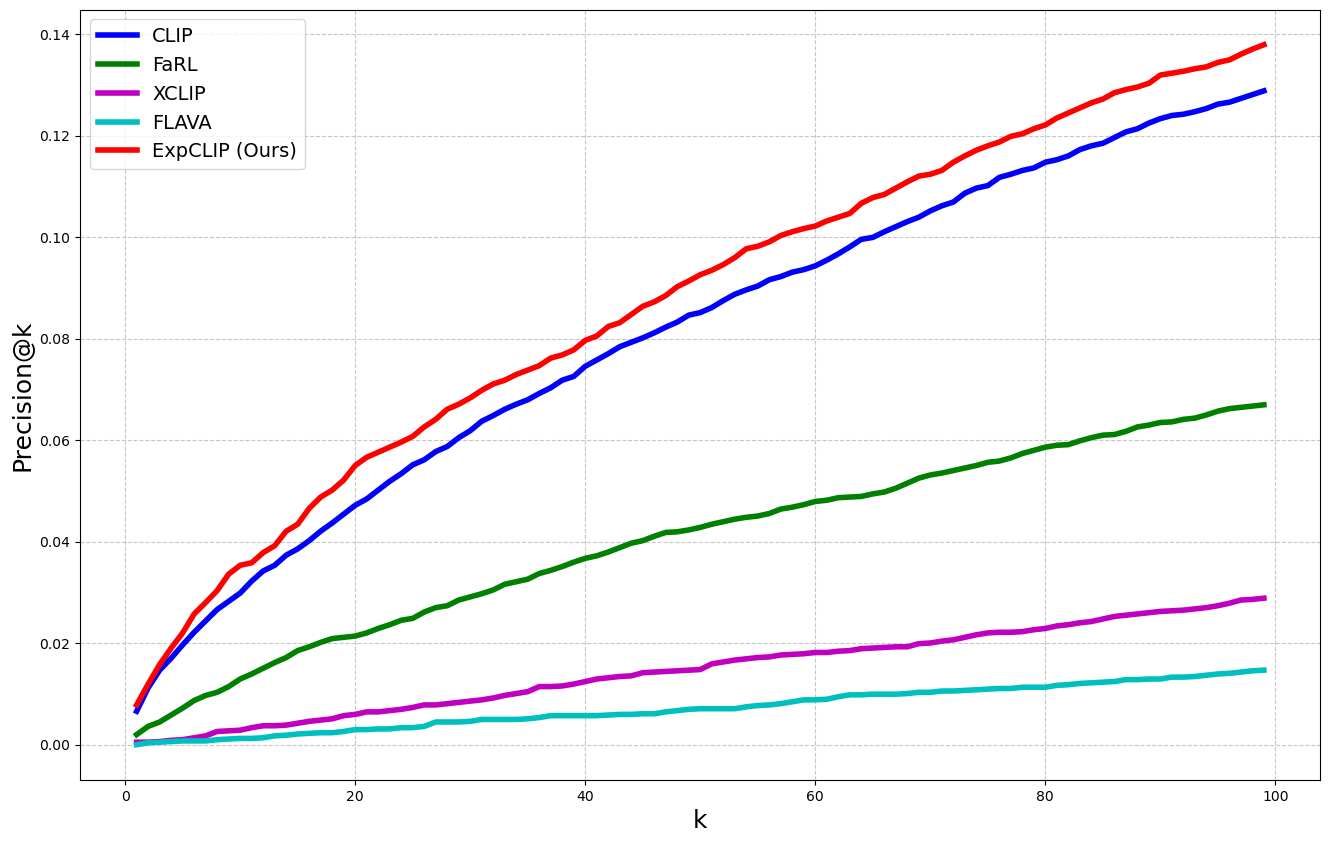}
	\caption{Comparison of the video-to-text retrieval performance on the subset of the MAFW.}
	\label{fig_D}
\end{figure}

\subsection{Normalized Feature Variance}
The normalized feature variance of each emotion on FERPlus is shown in Fig. \ref{fig_C}. The projected features demonstrate a lower variance than the CLIP feature space, proving the projected features are more concentrated.

\subsection{Evaluation of Vision-to-Text-Description Understanding Ability}
To verify the vision-to-text-description ability of our method, we conducted experiments using facial videos and their corresponding facial action descriptions. Specifically, we performed a video-text retrieval task on a subset of the MAFW dataset. This subset comprises 8,034 sample-level text descriptions, each detailing specific facial actions. We utilized Precision@k as the metric for video-to-text retrieval and compared the performance of our method against several other vision-language models. As illustrated in Fig. \ref{fig_D}, our Exp-CLIP model outperforms the other methods, which means our model retrieves more relevant text descriptions for a given facial video compared to other models and excels in understanding and retrieving complex and detailed facial action descriptions.

\subsection{More Sample-Level Predictions and Analysis}
\begin{figure*}[!t]
\renewcommand\thefigure{E}
	\centering
	\includegraphics[scale=0.8]{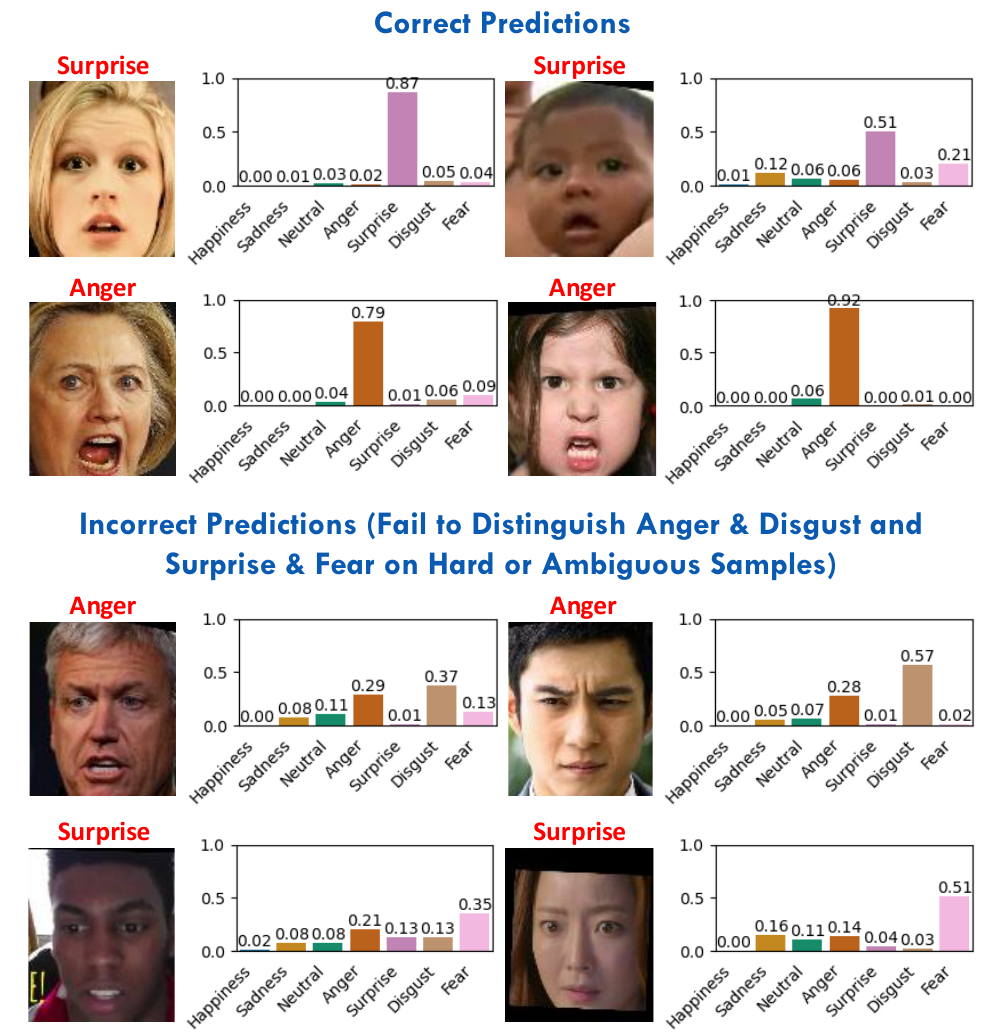}
	\caption{Zero-shot facial expression predictions of the proposed Exp-CLIP.}
	\label{fig_E}
\end{figure*}

We provide more sample-level predictions in the Fig. \ref{fig_E}. From the probability predictions on top images, we observe that our Exp-CLIP model correctly recognizes facial expressions with high confidence. However, our zero-shot method still makes some incorrect predictions, particularly when distinguishing between similar emotions like Anger and Disgust, as well as Surprise and Fear, especially on hard or ambiguous samples. These misclassifications likely arise from the subtle differences in facial expressions between these emotions, which can be difficult even for supervised models. Further refinement of our model or additional fine-tuning on few-shot labelled data may help mitigate these specific shortcomings.

\end{document}